\documentclass[10pt,journal,compsoc]{IEEEtran}

\usepackage{tabularx}
\usepackage{rotating}
\usepackage{graphicx}
\usepackage{amsmath}
\usepackage{amssymb}
\usepackage{cancel}
\usepackage{mdwlist}
\usepackage{subfig}
\usepackage{color}
\usepackage[ruled,vlined,noend]{algorithm2e}
\usepackage{hyperref}
\usepackage{ojw_style}

\usepackage{amsfonts}
\DeclareMathSymbol{\shortminus}{\mathbin}{AMSa}{"39}

\ifCLASSOPTIONcompsoc
\usepackage[nocompress]{cite}
\else
\usepackage{cite}
\fi

\begin{document}
%
\title{Least Squares Normalized Cross Correlation}

\author{Oliver J. Woodford
	\IEEEcompsocitemizethanks{\IEEEcompsocthanksitem Oliver J. Woodford was at Snap Inc.\ when this work was done.\protect\\
		E-mail: o.j.woodford.98@cantab.net}
}

%
%

\markboth{}%
{Woodford: Least Squares Normalized Cross Correlation}
%


\IEEEtitleabstractindextext{%
\begin{abstract}
	Direct methods are widely used for alignment of models to images, due to their accuracy, since they minimize errors in the domain of measurement noise. They have leveraged least squares minimizations, for simple, efficient, variational optimization, since the seminal 1981 work of Lucas \& Kanade, and normalized cross correlation (NCC), for robustness to intensity variations, since at least 1972. Despite the complementary benefits of these two well known methods, they have not been effectively combined to address \emph{local} variations in intensity. Many ad-hoc NCC frameworks, sub-optimal least squares methods and image transformation approaches have thus been proposed instead, each with their own limitations. This work shows that a least squares optimization of NCC without approximation is not only possible, but straightforward and efficient. A robust, locally normalized formulation is introduced to mitigate local intensity variations and partial occlusions. Finally, sparse features with oriented patches are proposed for further efficiency. The resulting framework is simple to implement, computationally efficient and robust to local intensity variations. It is evaluated on the image alignment problem, showing improvements in both convergence rate and computation time over existing lighting invariant methods.
\end{abstract}

}

\maketitle

\IEEEdisplaynontitleabstractindextext

%

\renewcommand{\paragraph}[1]{\par {\normalfont\sffamily\itshape #1:}~~}

\def\appendix{\appendices}

\renewcommand{\imI}[0]{\im{S}}
\renewcommand{\imJ}[0]{\im{T}}
\newcommand{\error}{\mathcal{E}}
\newcommand{\gain}{\alpha}
\newcommand{\bias}{\beta}
\newcommand{\warp}{\Pi}
\newcommand{\imtransform}{\Gamma}
\newcommand{\patchtransform}{\Psi}
\newcommand{\patch}{\bar{\im{I}}}
\newcommand{\patchI}{\bar{\imI}}
\newcommand{\patchJ}{\bar{\imJ}}
\newcommand{\patchzm}{\hat{\im{I}}}
\newcommand{\imsrc}{\imI}
\newcommand{\imtgt}{\imJ}
\newcommand{\Ix}{I_x}
\newcommand{\Iy}{I_y}
\newcommand{\graddir}{\vec{d}}
\newcommand{\gradscore}{s}
\newcommand{\patchcoords}{\vec{X}}
\newcommand{\patchcoordsx}{\vec{X}_i}
\newcommand{\patchcoordsY}{\vec{Y}}
\newcommand{\warpparams}{\vec{W}}
\newcommand{\warpparam}{\Delta}
\newcommand{\lieexp}[1]{e^{\hat{#1}}}
\newcommand{\lieexpsub}[2]{e^{\hat{#1}_{#2}}}
\newcommand{\deltawarp}{{\bf\Delta}}
\newcommand{\vzero}{\vec{0}}
\newcommand{\deltazero}{{\deltawarp}}
\newcommand{\warpdims}{\textrm{N}}
\newcommand{\spacedim}{\textrm{S}}
\newcommand{\patchlen}{\textrm{M}}
\newcommand{\numres}{\text{P}}
\newcommand{\optimparams}{\hat{\params}}
\newcommand{\robustify}{\rho}
\newcommand{\height}{\textrm{H}}
\newcommand{\width}{\textrm{W}}
\newcommand{\channels}{\textrm{C}}
\newcommand{\outchannels}{\textrm{D}}
\newcommand{\Mean}{\mean{\patch}}
\newcommand{\SD}{\sd{\patch}}
\newcommand{\SDzm}{\sd{\patchzm}}
\newcommand{\gb}{\mathrm{GB}}
\newcommand{\weightmatrix}{\im{F}}
\newcommand{\weight}{F}
\newcommand{\update}{\Phi}
\newcommand{\XinX}{\patchcoords\in\patches}
\newcommand{\jacX}{\jac_i}
\newcommand{\numpatches}{K}
\newcommand{\sumblocks}{\sum_{i=1}^\numpatches}
\newcommand{\patches}{\{\patchcoordsx\}_{i=1}^\numpatches}
\newcommand{\jacncc}{\jac_\patchtransform}
\newcommand{\M}{\mathbf{M}}
\newcommand{\g}{\mathbf{g}}
\newcommand{\offsets}{\vec{P}}
\newcommand{\coord}{\vec{x}}
\newcommand{\order}{o}

\newcommand{\patchIzm}{\hat{\imJ}}
\newcommand{\patchJzm}{\hat{\imI}}
\newcommand{\SDIzm}{\sd{\patchIzm}}
\newcommand{\SDJzm}{\sd{\patchJzm}}

\graphicspath{ {./fig/} }
\newlength{\imwidth}

\section{Introduction}
Aligning models to images has a huge variety of applications within Computer Vision, including alignment of satellite~\cite{barnea1972}, medical~\cite{hill2001medical} and surface metrology images~\cite{pan2013znssd}, object~\cite{stephens1990real} and keypoint~\cite{tomasi1991detection,wagner2009tracking} tracking, optical flow estimation~\cite{kroeger2016fast}, scene reconstruction~\cite{furukawa2010pmvs,schonberger2016colmapdense}, visual odometry~\cite{engel2017direct} and face model fitting~\cite{cootes2001aam,blanz2003morphable,matthews2004active}. These methods each seek to estimate a model which best explains the image data. Given that measurement noise in digital images manifests itself as errors in pixel intensity, probabilistic interpretations of such model fitting dictate that the most likely model is that which minimizes a photometric error between the model and the target image pixels. For this reason, ``direct'' methods, which minimize photometric errors, are widely used for model alignment tasks. However, one challenge remains poorly addressed: handling images with locally variable image intensities. These variations may be caused by scene effects such as non-Lambertian reflectance and a changing viewpoint, lighting effects such as shadows moving with the time of day, and camera effects such as lens vignetting, exposure time and image processing.
The ubiquity of these variations means that finding a photometric error and optimization framework that handles them effectively is highly consequential.

Direct methods date back to at least the early 1970s, with Barnea \& Silverman~\cite{barnea1972} using exhaustive search to find the optimal translation offset between two images. Higher dimensional problems require a more scalable solution, so in 1981 Lucas \& Kanade~\cite{lucas1981iterative} proposed an iterative optimization to compute relative camera pose and scene depth between two images. In particular, they used non-linear least squares optimization~\cite{tingleff2004methods}, a standard tool across many branches of science, due to the fact that Gauss' approximation to the Hessian makes second-order Newtonian optimization computationally efficient. Since then, inverse approaches have further improved the efficiency~\cite{hager1998efficient,baker2004lucas} and robustness~\cite{benhimane2004esm} of least squares methods on certain alignment problems, while applications of photometric least squares optimization have also increased, to cover the broad range of tasks mentioned above.

Lighting invariant direct methods have a similarly long history. Indeed, Barnea \& Silverman~\cite{barnea1972} used \emph{normalized cross correlation} (NCC) to provide invariance to a global change in gain and bias, while Lucas \& Kanade~\cite{lucas1981iterative} optimized over explicit gain and bias parameters for the same purpose. Since then, many methods have proposed new, iterative, NCC and intensity modelling approaches, as well as lighting invariant image transformations, to handle intensity changes between images. However, studying these approaches (in section \ref{sec:related_work}), we see that the complementary features of least-squares optimization, offering simple, efficient, gradient-based alignment, and the NCC score, offering robustness to lighting changes, have not yet been fully exploited. As a result, many of these other approaches, which are more computationally costly and have lower accuracy, see widespread use.

This work effectively combines least squares optimization and the NCC measure, for direct alignment that is robust to local intensity variations. The result is a simple to implement, efficient to optimize, photometric error that is robust under local lighting variations. Section \ref{sec:framework} presents the least squares NCC framework, including three key contributions:
\begin{enumerate*}
\item Deriving the Jacobian of the data normalization function used in least squares NCC, and an efficient method for computing it. This allows NCC to be optimized in a least squares framework, simply, efficiently, and without approximate derivatives. 
\item A locally normalized, robust formulation, which provides robustness to both \emph{local} intensity variations and occlusions.
\item Sparse, oriented patches with a novel sample layout, for improved efficiency.
 \end{enumerate*}
Section \ref{sec:evaluation} presents an evaluation of the framework against the current state of the art. The evaluation uses image alignment as the test problem. Pairs of image patches from different photos of the same surface taken in varying lighting conditions are aligned, and the resulting convergence rates, costs and optimization times are presented and discussed. Section \ref{sec:conclusion} concludes.

\section{Related work}
\label{sec:related_work}
This section discusses high level approaches to the direct alignment of images, covers the history of standard least squares photometric methods, enumerates three approaches to handling intensity variations within a least squares framework, then discusses previous NCC optimizers. 

\subsection{Image alignment}
Image alignment methods can be broadly split into direct and feature-based approaches. Feature-based approaches, such as that presented in Brown \& Lowe's Autostitch~\cite{brown2007autostitch}, pre-compute features in each image, match them across images, then minimize a geometric error on the correspondences.
This approach has the benefit of broad convergence, but is computationally expensive, and does not minimize the true measurement errors, which occur in the domain of pixel intensities, not feature locations.
Rather than requiring correspondences, direct methods minimize an error on pixel intensities or some transformed pixel space (see sec. \ref{sec:local_intensity}). These methods therefore exploit more image data, including edges, and have been shown to be more accurate than feature-based methods for direct odometry~\cite{engel2017direct}. Recent advances in efficient exhaustive search using fast fourier transforms and ridge regression~\cite{henriques2015kcf} have revolutionized \dims{2} tracking~\cite{vot2017results}, but do not extend to higher-dimensional alignment problems such as active appearance models~\cite{matthews2004active} and direct odometry~\cite{engel2017direct}. These latter problems use optimization-based approaches, which are efficient but require a good initialization. As NCC is a direct (photometric) error, and non-linear least squares is an optimization approach, further discussion is limited to direct, optimization-based methods.

\subsection{Least squares image alignment}
Non-linear least squares optimization frameworks~\cite{tingleff2004methods} are generally second order, Newtonian solvers that exploit Gauss' approximation to the Hessian, $\mat{H}\approx\jac^\trsp\jac$, where $\jac$ is the Jacobian of the residual errors whose squared sum is being minimized. This provides fast convergence (linear to quadratic), whilst only requiring the computation of the first derivatives of residuals, simplifying implementation and speeding up computation.

Lucas \& Kanade formulated and optimized a least squares photometric cost in 1981~\cite{lucas1981iterative}, applying their method to stereo image alignment. In the years since, many works have improved on the standard formulation in various ways, as chronicled in the ``Lucas-Kanade 20 years on'' series of articles~\cite{baker2004lucas,baker2003lucasp2,baker2003lucasp3}. These can be broadly categorized into modelling, efficiency, convergence and robustness improvements, which are described briefly below, as well as intensity variations, discussed in the following subsection.

\paragraph{Modelling} The original work~\cite{lucas1981iterative} sketched out approaches to both a \dims{2} transformation, in the form of an affine warp between image coordinates in two frames, and a \dims{3} transformation, optimizing scene depth and  a single, relative camera pose \wrt their photometric cost. Many works have followed with implementations of \dims{2} transformations, from translation only~\cite{tomasi1991detection}, through affine~\cite{bergen1992hierarchical} and  homographic~\cite{shum2001panorama} transformations, to quadratic~\cite{gao2015quadratic} ones most recently. Faces have been modelled with both \dims{2}~\cite{cootes2001aam} and \dims{3}~\cite{blanz2003morphable} deformable models, minimizing a transformation from an image to the model capturing variations in shape, appearance and lighting. Rigid \dims{3} alignment has also been applied to such tasks as object~\cite{hong20073d} and scene tracking~\cite{newcombe2011dtam}, plane fitting~\cite{habbecke2006plane,furukawa2010pmvs} and visual odometry~\cite{engel2017direct}. The methods proposed here are evaluated on \dims{2} homography alignment, but can be applied to all these tasks.

\paragraph{Efficiency}
Hager \& Belhumeur~\cite{hager1998efficient} first proposed an inverse additive update, which avoided the need to recompute the Jacobian of the least squares problem, dramatically improving the algorithm's efficiency. Baker \& Matthews~\cite{baker2001equivalence} then showed that an inverse compositional update could be used to apply the same trick to a wider variety of alignment parameterizations, excluding those with variable scene structure. Dellaert \& Collins~\cite{dellaert1999fast} improve efficiency by judiciously selecting a subset of pixels to optimize over.

\paragraph{Convergence}
Lucas \& Kanade~\cite{lucas1981iterative} recognized the small convergence basin of photometric costs, suggesting the now standard coarse-to-fine or multi-resolution strategy to counter this. Benihame \& Malis~\cite{benhimane2004esm} showed that combining forward and inverse Jacobians led to convergence more often, and in fewer iterations, than either the forwards or inverse approaches. Amberg \etal~\cite{amberg2009alignment} derived an update parameterization that better conditions the optimization problem, allowing first order optimization to run with similar speed but broader convergence than second order methods, when applied to active appearance models.

\paragraph{Robustness}
The use of robust functions to reduce the impact of outliers in least squares costs is well known~\cite{black1996robust}. Baker \& Matthews~\cite{baker2003lucasp2} describe the application of such functions to direct alignment, and their optimization using \emph{iteratively reweighted least squares} (IRLS)~\cite{holland1977irls}.

The methods proposed here can benefit from all these improvements in efficiency, convergence and robustness.

\subsection{Handling intensity variations}
\label{sec:local_intensity}
Three main categories of direct alignment methods currently exist to handle data with local intensity variations. We enumerate these with regard to image-based alignments, where scene effects such as non-Lambertian reflectance and a changing viewpoint, lighting effects such as shadows moving with the time of day, and camera effects such as lens vignetting, all create local changes in intensity.

\paragraph{Intensity models} The first category of methods estimate parameters of an intensity model. Lucas \& Kanade~\cite{lucas1981iterative} originally proposed a two-parameter gain and bias model, which can only model a global change in intensity. Silveira \& Malis~\cite{silveira2007real} extend the gain model to independent regions, but retain the global bias. This allows for more local intensity variations, at the expense of a higher-dimensional search space. Efficient inverse compositional optimization methods for such costs have also been developed~\cite{bartoli2008generative,li2018photometriclie}. More complex image formation models optimize albedo and reflectance, as well as shape \cite{haefner2019iccv}. A deficiency of such approaches is that they introduce additional variables to be optimized, slowing computation. Even assuming constant brightness, varying camera exposures change the pixel values of a given object. Some recent visual odometry methods model exposure time per frame \cite{engel2017direct,bergmann2018exposure}, but the underlying assumption remains constraining.

\paragraph{Invariant measures} The second category consists of lighting invariant scores between aligned image patches. This category includes NCC, the subject of this paper, as well as mutual information~\cite{viola1997mutualinformation,maes1997multualinformation,dame2010mutualinfo}. Mutual information supports a wider range of lighting changes than the affine-invariance of NCC, but nevertheless requires a globally consistent transformation of intensities. In addition, a least squares formulation does not exist. This category retains both the rotational invariance and the dimensionality of the SSD formulation, has no precomputation, and in the case of NCC introduces only a modest computational overhead per iteration.

\paragraph{Image transformation} The final category is image transformation. Prior to least squares optimization, images are converted pixelwise into a lighting invariant space. Some methods compute distance transforms over edge images~\cite{kuse2016edgetracking,wang2016edge}, converting intensity into a form of geometric, rather than photometric, error. Other methods compute a multi-channel descriptor (of 4-36 channels) per pixel, based on the local image texture, some of which are hand designed~\cite{hirschmuller2008evaluation,crivellaro2014descriptorfields,antonakos2015feature,alismail2016robust}, others of which are learned~\cite{chang2017clkn,vonstumberg2020gn}. This pre-process step increases computation time, but furthermore, as noted in section \ref{sec:feature_selection}, alignment computation time also increases linearly with the number of image channels. These descriptors are often not invariant to in-plane image rotations,\!\!\footnote{Invariance of a pixelwise transformation $T(\cdot)$ to an in-plane rotation $R(\cdot)$ requires $T(R(\im{I}))= R(T(\im{I}))$, where $\im{I}$ is an image.} so alignment in such cases will fail. However, it is also possible to apply these transformations after, instead of before, warping, shifting such approaches to the second category, invariant measures
. Transformations applied after warping are invariant to in-plane rotations, but incur the computational cost of transformation at every iteration, rather than just once. There is some disagreement on whether image transformations result in higher accuracy when applied before or after warping; some~\cite{antonakos2015feature} conclude the former is better, others~\cite{alismail2016robust} the latter.

Several studies compare subsets of these approaches across various alignment tasks with intensity changes, including stereo~\cite{hirschmuller2008evaluation}, optical flow~\cite{vogel2013evaluation}, homographic alignment~\cite{alismail2016robust}, direct visual odometry~\cite{park2017illumination} and active appearance models~\cite{antonakos2015feature}. Four of these~\cite{alismail2016robust,hirschmuller2008evaluation,park2017illumination,vogel2013evaluation} conclude the 8-channel census image transform performs amongst the best, while another, studying a different set of image transforms~\cite{antonakos2015feature}, cites the 36-channel HoG descriptor~\cite{dalal2005hog}.

\subsection{Optimization of NCC}
\label{sec:previous_ncc}
A common approach to NCC optimization when solving over one or two parameters, such as finding the translation or depth of an image patch, is to employ an exhaustive search~\cite{vogiatzis2011video,wagner2009tracking,sun2010ncc}. This is often used in production keypoint tracking systems for SLAM~\cite{wagner2009tracking} and real-time dense stereo methods~\cite{vogiatzis2011video}, as it has a wider convergence basin, but it does not scale well to higher dimensional search spaces. Two popular multi-view stereo pipelines~\cite{furukawa2010pmvs,schonberger2016colmapdense} that optimize NCC over depth and patch normal use BFGS~\cite[\S6.1]{nocedal2006optimization} with numerical gradients~\cite{furukawa2010pmvs} (computed using finite differences), or a sampling approach~\cite{schonberger2016colmapdense}, both of which are inefficient.

Gradient-based methods, which include least squares optimization, scale well with dimensionality by traversing the state space one direction (of high gradient) at a time. NCC has previously been optimized using gradient-based methods with analytic gradients, particularly for image alignment: Irani \& Anandan~\cite{irani1998robust} use a second-order Newtonian framework, requiring the computation of second derivatives, which increases computational cost and amplifies image noise. They handle local intensity variations by maximizing a sum of NCC scores over subregions. Brooks \& Arbel~\cite{brooks2006generalizing} extend inverse compositional optimization to arbitrary cost functions (including NCC), using the BFGS~\cite[\S6.1]{nocedal2006optimization} optimizer: a first order optimizer which builds an approximation to the Hessian over time. Evangelidis \& Psarakis~\cite{evangelidis2008parametric} compute an iterative update, derived by passing a Taylor expansion of a warped image patch through the NCC measure, in an algorithm they call \emph{enhanced correlation coefficient maximization} (ECCM). This method essentially constructs the Jacobian of the zero-mean patch, and a function that is minimized with respect to a scalar parameter, $\lambda$, which then scales the zero-mean patch intensity, to account for the difference in normalization between patches. Issues with this approach are that the computation of $\lambda$ is non-negligible, especially in the efficient inverse compositional case, and furthermore that it does not extend easily to normalizing multiple regions simultaneously, nor robustifying such costs. These challenges are highlighted in an extension to \dims{2} alignments on pixelwise normalized image gradients~\cite{lamprinou2016robust}, that requires an even more bespoke framework, in which the warp parameters are jointly optimized with all $\lambda$s, one per pixel. Scandaroli \etal~\cite{scandaroli2012ncc} derive an approximate Newton update using only first derivatives, which is shown (Appendix \ref{sec:scandaroli}) to be equivalent to the simpler least squares update derived here, also locally normalizing subregions. Both Newton methods~\cite{irani1998robust,scandaroli2012ncc} weight the influence of subregions on the update, to improve robustness to outliers. The schemes used to compute weights are ad-hoc, the former using the determinant of the subregion hessian, the latter using a k-means clustering of NCC scores, combined with the Huber kernel.

Two methods using an NCC equivalent cost in a least squares framework have previously been proposed~\cite{baker2003lucasp3,pan2013znssd}. Baker \etal~\cite{baker2003lucasp3} present the Normalized Inverse Compositional algorithm, while in the field of surface metrology, where direct alignment is known as digital image correlation~\cite{wang2015compare}, Pan \etal~\cite{pan2013znssd} propose the ZNSSD framework. Both methods employ the least squares NCC error used here (\eqnref{error}), within a standard least squares framework, allowing the frameworks to exploit the inverse compositional method for efficiency, and robust kernels for robustness. However, both make the same approximation to the Jacobian of the error, replacing the derivative of patch normalization with the identity matrix. The Jacobian is therefore slightly inaccurate, generally slowing and occasionally impeding convergence of the optimization. The two methods handle normalization in the error slightly differently also: The latter approach~\cite{pan2013znssd} uses the difference between normalized samples (\ie the true error), whereas the former~\cite{baker2003lucasp3} projects out differences in gain and bias from the error, with respect to some basis of projection (\ie an approximation of the error). The framework presented here is to my knowledge the first that computes the exact Jacobian of the least squares NCC error (\eqnref{jac_ncc}). When using a novel, locally normalized formulation, capable of handling local intensity variations, the exact Jacobian confers faster and more frequent convergence, as shown in section \ref{sec:optim_comp}.

This abundance of methods demonstrates one thing: despite its broad use, there is a lack of consensus on the best way to optimize NCC. This paper addresses that issue.

\section{Least squares NCC framework}
\label{sec:framework}
\subsection{Least squares NCC cost}
\label{sec:formulation}
Let us define a normalization function, $\patchtransform : \real^\patchlen \rightarrow \real^\patchlen$, which takes a vector of length $\patchlen$, and normalizes it such that it has zero mean and unit variance, thus:
\begin{align}
\label{eqn:patch_normalize}
\patchtransform(\patch) &= \frac{\patch - \Mean}{\SD},~~~~~
\Mean = \frac{\onestrsp\,\patch}{\patchlen},~~~~~
\SD = \|\patch - \Mean\|,
\end{align}
where $\ones$ is a vector of ones.\!\!\footnote{Similarly, $\zeros$ is a vector of zeros.}
The standard formulation of NCC~\cite[eqn.~8.11]{szeliski2010computer},  a measure of similarity between a source and target vector, $\patchI$ and $\patchJ$ respectively, which should be \emph{maximized}, can then be written as
\begin{align}
\label{eqn:ncc_standard}
\E{NCC}(\patchI, \patchJ) = \patchtransform(\patchI)^\trsp \patchtransform(\patchJ).
\end{align}
This measure is bounded, in the range $[-1,1]$.
It is known~\cite{evangelidis2008parametric,pan2013znssd} that this is equivalent (negated, up to scale and ignoring a scalar offset) to the sum of squared differences between the two normalized vectors, thus:
\begin{align}
\label{eqn:ncc_least_squares}
\E{ZNSSD}(\patchI, \patchJ) =& \|\patchtransform(\patchI)-\patchtransform(\patchJ)\|^2,\\
=& \|\patchtransform(\patchI)\|^2 + \|\patchtransform(\patchJ)\|^2 - 2 \patchtransform(\patchI)^\trsp \patchtransform(\patchJ),\\
\label{eqn:ncc_cost_identity}
=&~2 - 2\E{NCC}(\patchI, \patchJ),
\end{align}
since $\|\patchtransform(\cdot)\|^2=1$ by definition. This latter measure, which is \emph{minimized}, not only has its optimum at the same inputs, but the gradients of the two measures are proportional (with a scale factor of -2) for all inputs. It is bounded in the range $[0, 4]$. Importantly, since it is formulated as a sum of squared errors, it is readily optimized within a least squares framework~\cite{tingleff2004methods}. This confers the benefit that between linear and quadratic convergence of the measure can be achieved using only first derivatives of the error, $\patchtransform(\patchI)-\patchtransform(\patchJ)$. Let us therefore refer to $\E{ZNSSD}$ as the \emph{least squares NCC cost}.

\subsection{Alignment error}
NCC is generally optimized with respect to an alignment between source and target data samples, $\imsrc$ and $\imtgt$, that are scalar or vector fields of $n$ dimensions; the alignment is found such that the aligned samples are most similar, under the measure. These fields could be \dims{1} (\eg audio data),  \dims{2} (\eg image data), \dims{3} (\eg volumetric data), or higher. This work applies the formulation to single channel image data, therefore this problem is assumed henceforth, to simplify notation. However, note that the contributions of this work can easily be applied to other dimensionalities of data, and that least squares trivially generalizes to multiple channels~\cite{alismail2016robust}, \ie vector fields.

Source and target here are therefore images, $\imsrc, \imtgt \in \real^{\height\times\width}$. The coordinates of values within the target image to be aligned are given by matrix of $\patchlen$ homogeneous coordinates, denoted $\patchcoords \in \real^{\spacedim \times \patchlen}$, where $\spacedim \ge 3$. Often $\patchcoords$ consists of the dense grid of all pixel coordinates in $\imtgt$, but this is by no means required; the coordinates can be arbitrary, a fact exploited here by using a non-grid based selection of points, described in section \ref{sec:feature_selection}. These coordinates can be of greater than 3 dimensions if, for example, they represent the \dims{3} location of the surface visible in the image (where $\spacedim = 4$), or a quadratic alignment is used~\cite{gao2015quadratic} (where $\spacedim = 6$). The transformation of coordinates $\patchcoords$ from the target to source coordinate frame is given by
\begin{align}
\label{eqn:warp}
\patchcoordsY &= \proj(\warpparams \patchcoords),
\end{align}
where $\warpparams \in \real^{3\times\spacedim}$ is the warp matrix, which applies a linear transformation to the coordinates, and $\proj : \real^{3\times\patchlen}\rightarrow\real^{2\times\patchlen}$ applies any non-linearities present in the measurement process to each column of $\warpparams\patchcoords$; in the case of images, this is a projection onto the image plane, and correction for camera calibration and lens distortion. The evaluation tasks used here require only a simple projection: $\proj([x,y,z]^\trsp) = [x/z, y/z]^\trsp$. Other data modalities, and indeed some image warps (\eg an affine warp), do not require a projection, in which case $\proj$ is the identity function. The warp matrix $\warpparams$ may over-parametrize the transformation, but is kept on the manifold of allowed transformations using a special update parametrization (see sec. \ref{sec:warp_update}). The source data is sampled at the coordinates $\patchcoordsY$, using interpolation,\!\!\footnote{Any differentiable interpolation scheme can be used. Linear interpolation (bilinear, for images) is used here.} producing a sampled data vector, denoted by a bar, \eg $\patch$. This sampling is expressed using the notation
$\patch = \im{I}(\patchcoordsY)$.

The entire warp process described above is denoted by the warp function $ \warp : \real^{\height\times\width}, \real^{\spacedim\times\patchlen}, \real^{3\times\spacedim}\rightarrow \real^{\patchlen}$, thus:
\begin{align}
\label{eqn:warp_func}
\patch &= \warp(\im{I}, \patchcoords, \warpparams),\\
&= \im{I}( \proj(\warpparams \patchcoords)).
\end{align}
The least squares NCC error for the alignment between source and target is then given by
\begin{align}
\label{eqn:error}
\error_{\imsrc,\imtgt,\patchcoords}(\warpparams) &= \patchtransform(\warp(\imsrc, \patchcoords, \warpparams)) - \patchtransform(\warp(\imtgt, \patchcoords, \identity)),
\end{align}
where $\identity$ is the identity matrix.

\subsection{Cost functions}
Using the notation introduced thus far, the standard least squares NCC cost function is thus
\begin{align}
\label{eqn:global_cost}
\E{global} &= \| \error_{\imsrc,\imtgt,\patchcoords}(\warpparams) \|^2.
\end{align}
This is referred to as the global cost, because it is invariant to a global gain and bias. Similar to local subregion NCC in some earlier works~\cite{irani1998robust,scandaroli2012ncc}, a cost function consisting of a sum of local, least squares NCC costs is introduced:
\begin{align}
\label{eqn:local_cost}
\E{local} &= \sumblocks\robustify \big( \| \error_{\imsrc,\imtgt,\patchcoordsx}(\warpparams) \|^2),
\end{align}
where $\{\patchcoordsx\}_{i=1}^\numpatches$ are the sample coordinates of a set of $\numpatches$ local patches (described in sec.\ \ref{sec:feature_selection}), making this cost invariant to local variations in intensity. Additionally, each of the summed costs is robustified by a function\footnote{Suitable robustification functions satisfy $\robustify(0)=0$ and 
increase monotonically for positive input values. Note that robustifying a single cost term, such as \eqnref{global_cost}, is redundant. Since $\robustify$ increases monotonically, the minima occur at identical warps.} $\robustify : \realpos \rightarrow \realpos$, that downweights large errors. Though the least squares NCC cost is bounded, 
as with SSD it is rare that large costs, especially those $>2$, which indicate inversely correlated patches, provide a gradient towards the global optimum, in contrast to costs close to zero. Yet in a sum of such costs, larger costs will have a greater influence on the step taken. Robustification ensures that costs close to converging have more influence than costs far from converged.

This work uses the Geman McClure robustifier~\cite{black1996robust,geman1985bayesian},
\begin{equation}
\label{eqn:geman_mcclure}
	\robustify(s) = \frac{s}{s + \tau^2},
\end{equation}
where $\tau = 0.5$ here. This kernel is fully truncated, since $\displaystyle \lim_{s\to\infty} \robustify(s) = 1$, such that gross outliers do not impact the optimal warp parameters, but also tends to zero gradient only in the limit ($s\to\infty$), ensuring broad convergence.

The robustification of costs to outlier data, such as occluded regions, is a standard tool in vision literature~\cite{black1996robust,triggs1999bundle}, and has often been applied, pixelwise, to least squares image alignment using the SSD cost~\cite{baker2003lucasp2,engel2017direct}. Whilst pixelwise robustification of a least squares NCC cost is possible, outlier samples will affect the normalization of the entire patch, affecting even non-outlier samples, therefore it is less appropriate. Robustification is therefore applied to each normalized patch as a whole; this may be novel in the domain of least squares image alignment, but is a standard approach in broader least squares applications, \eg bundle adjustment~\cite{triggs1999bundle}, and software tools, \eg ceres solver~\cite{ceres-solver}, which does not affect optimization. In addition, the downweighting of NCC scores over local image regions has previously been proposed in other optimization frameworks~\cite{irani1998robust,scandaroli2012ncc}. 

The impact of using a sum of costs, and robustifying those costs, are evaluated separately in section \ref{sec:evaluation}, where the impact of using a sparse set of judiciously chosen sample points (described in sec. \ref{sec:feature_selection}) is also evaluated.

\subsection{Optimization}
The presented cost functions can be optimized using a non-linear least squares solver, which involves parametrizing updates to the variables, computing the cost function and its derivatives with respect to the update variables, and iterating an update rule until convergence. These steps are described below.

\subsubsection{Warp update parametrization}
\label{sec:warp_update}
The manifold of allowed warp transformations has an intrinsic dimensionality, $\warpdims$, which may be less than the dimensionality of the warp matrix. A special update, parametrized by an update vector $\deltawarp \in \real^{\warpdims}$, is therefore required to keep the warp matrix on the manifold.
Given such an update vector, which is computed at each iteration of the optimization (see below), the warp is updated as follows:
\begin{align}
\warpparams &\leftarrow \warpparams\update(\deltawarp),
\end{align}
where $\update : \real^{\warpdims} \rightarrow \real^{\spacedim\times\spacedim}$ converts an update vector into a warp matrix, such that the set of warps is a group~\cite{baker2004lucas}. In particular, $\update(\vzero)=\identity$, and $\lim_{\deltawarp\rightarrow\vzero}\update^{-1}(\deltawarp) = \update(-\deltawarp)$.
The update must be applied via a composition to the \emph{right} of the warp in order to support inverse~\cite{baker2004lucas} and ESM~\cite{benhimane2004esm} approaches (sec. \ref{sec:jacobian}). The  transformation of target data coordinates into the target, via the source, can then be written as $\left(\warpparams\update(\deltawarp)\right)^{-1}\warpparams = \update(-\deltawarp)\warpparams^{-1}\warpparams = \update(-\deltawarp)$,\!\!\footnote{A left update results in $(\update(\deltawarp)\warpparams)^{-1}\warpparams \ne \update(-\deltawarp)$, while $\update(\deltawarp) \warpparams\warpparams^{-1} = \update(\deltawarp)$ is a transformation of \emph{source} data coordinates into the source, via the target. If $\spacedim>3$, the matrix inverse is replaced with the pseudo-inverse.} which allows for an alternative parametrization of the target (rather than source) data, $\warp(\imtgt, \patchcoords, \update(-\deltawarp)))$. A right update is also required when $\spacedim>3$.

This work uses warp update parametrization
\begin{align}
\label{eqn:update_homog}
\update(\deltawarp) &= \left[\begin{array}{ccc} 1+\warpparam_4+\warpparam_5 &\warpparam_6-\warpparam_3 &\warpparam_1\\ \warpparam_6+\warpparam_3 &1+\warpparam_4-\warpparam_5 &\warpparam_2\\ \warpparam_7& \warpparam_8 &1-2\warpparam_4\end{array}\right],
\end{align}
based on generators of the SL(3) Lie group~\cite{benhimane2004esm}\cite[eqn. 85--87]{eade2014liegroups}. Parameters $\warpparam_1$ \& $\warpparam_2$ encode translation, $\warpparam_3$ (in-plane) rotation, $\warpparam_4$ scale, $\warpparam_5$ \& $\warpparam_6$ complete the affine warp, and $\warpparam_7$ \& $\warpparam_8$ add perspective foreshortening for a full homography; lower dimensional parametrizations can thus be derived by setting unused parameters to zero.

\subsubsection{Iterative update}
\label{sec:gauss_newton}
Any standard non-linear least squares optimizer (\eg Gauss-Newton, Levenberg-Marquardt, \etc~\cite{tingleff2004methods}) can be used to optimize the costs given in equations (\ref{eqn:global_cost}) \& (\ref{eqn:local_cost}), by iteratively minimizing a linear least squares approximation to the non-linear cost. Here Gauss-Newton is used, such that the per iteration update for equation (\ref{eqn:local_cost}) (of which equation (\ref{eqn:global_cost}) is a special case) is computed thus:
\begin{align}
\label{eqn:linear_system}
\deltawarp &= \argmin_{\deltawarp^{\!\!*}}\|\error+\jac\deltawarp^{\!\!*}\|^2 ~= -\argmin_{\deltawarp^{\!\!*}}\|\jac\deltawarp^{\!\!*}-\error\|^2\\ 
\error &= \left[\begin{array}{c}\sqrt{\robustify'_1}\error_{\imsrc,\imtgt,\patchcoords_1}(\warpparams)\\\vdots \\ \sqrt{\robustify'_\numpatches}\error_{\imsrc,\imtgt,\patchcoords_\numpatches}(\warpparams)\end{array}\right], ~~~~\jac=\left[\begin{array}{c}\sqrt{\robustify'_1}\jac_1\\\vdots \\ \sqrt{\robustify'_\numpatches}\jac_\numpatches\end{array}\right] \\
\label{eqn:patch_jacobian}
\jacX &= \frac{\partial}{\partial \deltawarp}\error_{\imsrc,\imtgt,\patchcoordsx}(\warpparams\update(\deltazero))\Bigr|_{\deltawarp=\vzero},\\
&= \left[\frac{\partial}{\partial \warpparam_1}\error_{\imsrc,\imtgt,\patchcoordsx}(\cdot)  ~\dots~  \frac{\partial}{\partial \warpparam_\warpdims}\error_{\imsrc,\imtgt,\patchcoordsx}(\cdot)\right],\\
\robustify'_i &= \frac{\partial}{\partial s}\robustify(s)|_{s=\|\error_{\imsrc,\imtgt,\patchcoordsx}(\warpparams)\|^2}.
\end{align}
The scalar values $\sqrt{\robustify'_i}$ are weights which account for robustification, in a scheme known as iteratively reweighted least squares~\cite{holland1977irls}. The weighting serves a similar purpose to the ad-hoc weighting of NCC scores from subregions employed in previous works~\cite{irani1998robust,scandaroli2012ncc}, but with a rational origin in the cost function. Details on how the linear least squares solution to equation (\ref{eqn:linear_system}) is computed are given in section \ref{sec:solve_linear}. This step is repeated until one of the convergence criteria, detailed in section \ref{sec:convergence_criteria}, are met.

\subsubsection{Jacobian computation}
\label{sec:jacobian}
The following three approaches to Jacobian computation for compositional warp updates have been proposed in the literature.
\paragraph{Forwards compositional} The standard Jacobian~\cite{baker2004lucas} is given by a straightforward differentiation of equation (\ref{eqn:error}): $\overrightarrow{\jac} = \frac{\partial}{\partial \deltawarp}\patchtransform(\warp(\imsrc, \patchcoords, \warpparams\update(\deltazero)))$.
\paragraph{Inverse compositional} Jacobians can also be computed in the target image, at the identity warp~\cite{baker2004lucas,hager1998efficient}, $\overleftarrow{\jac} = -\frac{\partial}{\partial \deltawarp}\patchtransform(\warp(\imtgt, \patchcoords, \update(-\deltazero)))$,\!\!\footnote{Note that the two minus signs cancel each other out.} such that they are constant. When Gauss-Newton is used, the pseudo inverse $\jac^+ = (\jac^\trsp \jac)^{-1} \jac^\trsp$ can also be precomputed, resulting in a much faster update. However, robust kernels weight the Jacobian, therefore change the value of the pseudo inverse at each iteration, so it cannot be precomputed when robustification is used. 
\paragraph{Efficient Second-order Minimization (ESM)} Taking the average of the above two Jacobians,  $\overleftrightarrow{\jac} = \frac12 \left(\overleftarrow{\jac} + \overrightarrow{\jac}\right)$, has been shown to improve both the rate (number successful) and speed (number of iterations) of convergence~\cite{benhimane2004esm}.

\subsubsection{NCC Jacobian}
\label{sec:ncc_derivative}
As mentioned earlier, no previous work (to my knowledge) has computed the Jacobian of the least squares NCC cost function, $\patchtransform$, therefore they have not computed the true least squares update. However, it should be noted that all parts of \eqnref{patch_normalize} are differentiable. 
The analytic Jacobian $\jacncc=\frac{\partial}{\partial\patch}\patchtransform(\patch)$
can therefore be computed 
first by subtracting the mean, then by applying the variance normalization using the quotient rule, as follows: 
\begin{align}
\label{eqn:jac_zero_mean}
\frac{\partial}{\partial\patch}(\patch-\Mean) &= \identity - \frac{\ones\onestrsp}{\patchlen},\\
\frac{\partial}{\partial\patchzm}\frac{\patchzm}{\SDzm} &= \frac{\identity\SDzm - \patchzm^\trsp\patchzm/\SDzm}{\SDzm^2},
\end{align}
where $\patchzm=\patch-\Mean$. Applying the chain rule and some rearrangement (using $\patchtransform(\patch)=\patchzm/\SDzm$), this gives
\begin{align}
\label{eqn:jac_ncc}
\jacncc &= \underbrace{\frac{\identity - \patchtransform(\patch)\patchtransform(\patch)^\trsp}{\SD}}_\textrm{Variance norm.}\underbrace{\left(\identity - \frac{\ones\onestrsp}{\patchlen}\right)}_\textrm{Zero mean}.
\end{align}
Presenting and applying this formula is the first contribution of this work. It can be shown analytically that the resulting update step is equivalent (for forward and inverse compositional schemes, but not ESM) to the step of Scandaroli \etal~\cite{scandaroli2012ncc} (see Appendix \ref{sec:scandaroli}). However, the application of this formula is simpler, plugging straight in to existing photometric least squares frameworks, and benefits from the substantial body of literature in that area, as results using ESM and a standard robustifier show (in section \ref{sec:optim_comp}).

Using the chain rule, the full Jacobian of \eqnref{patch_jacobian} is given by $\jacncc\jac_{\warp}$, the latter term representing the $\patchlen\times\warpdims$ Jacobian of $\warp$, $\frac{\partial}{\partial\deltawarp}\warp(\im{I},\patchcoords,\warpparams\update(\deltawarp))$. The warp-parameterization-dependent formulae for $\jac_{\warp}$ are not presented here. These are available in many prior works, \eg Baker \& Matthews~\cite{baker2004lucas}. Also note that modern auto-differentiation tools, such as that included in ceres solver~\cite{ceres-solver}, can automatically compute analytic derivatives at runtime.

\subsubsection{Efficient  Jacobian computation}
\label{sec:ncc_three_pass}
Two tricks can be used to significantly speed up computation of the Jacobian, $\jacX$. The first is to note that while $\jacncc$ is $\patchlen\times\patchlen$, it is two rank-one updates to $\jac_{\warp}$, and can therefore be computed as
\begin{align}
\jacncc\jac_{\warp} &= \frac1{\SD} \left(\hat\jac - \patchtransform(\patch)\left(\patchtransform(\patch)^\trsp \hat\jac\right)\right), \\ 
\hat\jac &= \jac_{\warp} - \frac{\ones}{\patchlen}\left(\ones^\trsp\jac_{\warp}\right).
\end{align}
This reduces the computation of the product $\jacncc\jac_{\warp}$ from a na\"ive $\bigO(\patchlen^2)$ to a linear $\bigO(\patchlen)$.

A further speed up can be achieved if $\warpdims > 2$, by noting that the complete Jacobian can be further broken down into $\jacncc\jac_{\im{I}}\jac_{\proj}$, where $\jac_{\proj}$ is the $2\patchlen\times\warpdims$ Jacobian of target to source coordinate transformation, and $\jac_{\im{I}}$ is the $\patchlen\times2\patchlen$ block-diagonal Jacobian of image gradients at the target sample points. Computing $\jacncc(\jac_{\im{I}}\jac_{\proj})$ requires approximately $5\warpdims$ multiplications per sample, whereas computing $(\jacncc\jac_{\im{I}})\jac_{\proj}$ requires just $2\warpdims + 6$. For an $\dims{8}$ homography parametrization, this results in $\sim$45\% less computation.

\subsection{Acceleration through sample sparsity}
\label{sec:feature_selection}
The computational complexity of the direct least squares methods is determined by three main steps: The computation of pixelwise errors and the Jacobian of these errors, the construction of the linear system, and solving the linear system. Letting $\numres$ denote the number of errors, which, in the case of image alignment, is a product of the numbers of patches, samples per patch and data channels, and recalling that $\warpdims$ is the dimensionality of the transformation, the complexity is given by:
\begin{align}
\label{eqn:complexity}
\underbrace{\bigO(\numres\warpdims)}_{\text{Comp. $\jac$}} + \underbrace{\bigO(\numres\warpdims^2)}_{\jac^\trsp\jac} + \underbrace{\bigO(\warpdims^3)}_{\text{Lin. solve}}.
\end{align}
Generally $\warpdims$ is determined by the problem to be solved, thus fixed, with $\warpdims \ll \numres$. Computation can therefore only be reduced by reducing $\numres$, on which it depends linearly.

Unlike image transformation methods that increase $\numres$ with more image channels~\cite{crivellaro2014descriptorfields,antonakos2015feature,alismail2016robust,chang2017clkn,vonstumberg2020gn}, NCC is applied directly to the original data, or, as here, a single channel, grayscale version of the data, \ie the number of channels is \emph{reduced}, thus already conferring a computational advantage. However, this can be further improved by reducing the number of points sampled over.

\paragraph{Per-pixel grid samples} In standard approaches, the set of sample points used ($\patchcoords$ in $\E{global}$ and $\{\patchcoordsx\}_{i=1}^\numpatches$ in $\E{local}$) is usually a grid of points, one per pixel (in the case of images). In this work grid samples are placed on pixel corners, since under bilinear interpolation a sample must shift at least half a pixel in any direction before the image intensity changes non-linearly, in contrast to pixel centers, where the intensity is non-differentiable. In the case of $\E{local}$, each distinct $N\times N$ block of samples in this grid is locally normalized, where $N=6$ for most experiments, but a range of values is also evaluated in section \ref{sec:block_size_evaluation}.

\paragraph{Sparse samples} It has previously been proposed~\cite{dellaert1999fast} to select a subset of sample points, reducing $\numres$ and hence computation time, judiciously, such that accuracy is not impacted. In that work, points are chosen based on evaluating Jacobians for every pixel, which for a real-time system might be prohibitively expensive, even when computed only once, on the first frame. More recently, sparse patches have been proposed for direct visual odometry~\cite{engel2017direct}, using image gradient magnitude as a patch selection score. This work also selects patches using image gradient magnitude, but proposes an oriented patch layout that is shown (sec.\ \ref{sec:error_comp}) to improve convergence.

\subsubsection{Sparse patch selection}
\begin{figure}[t]
	\setlength{\imwidth}{0.33\linewidth}
	\begin{tabularx}{\linewidth}{@{}*{2}{c@{\extracolsep{\fill}}}c@{}}
		\includegraphics[width=\imwidth]{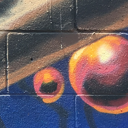}&
		\includegraphics[width=\imwidth]{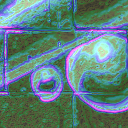} &
		\includegraphics[width=\imwidth]{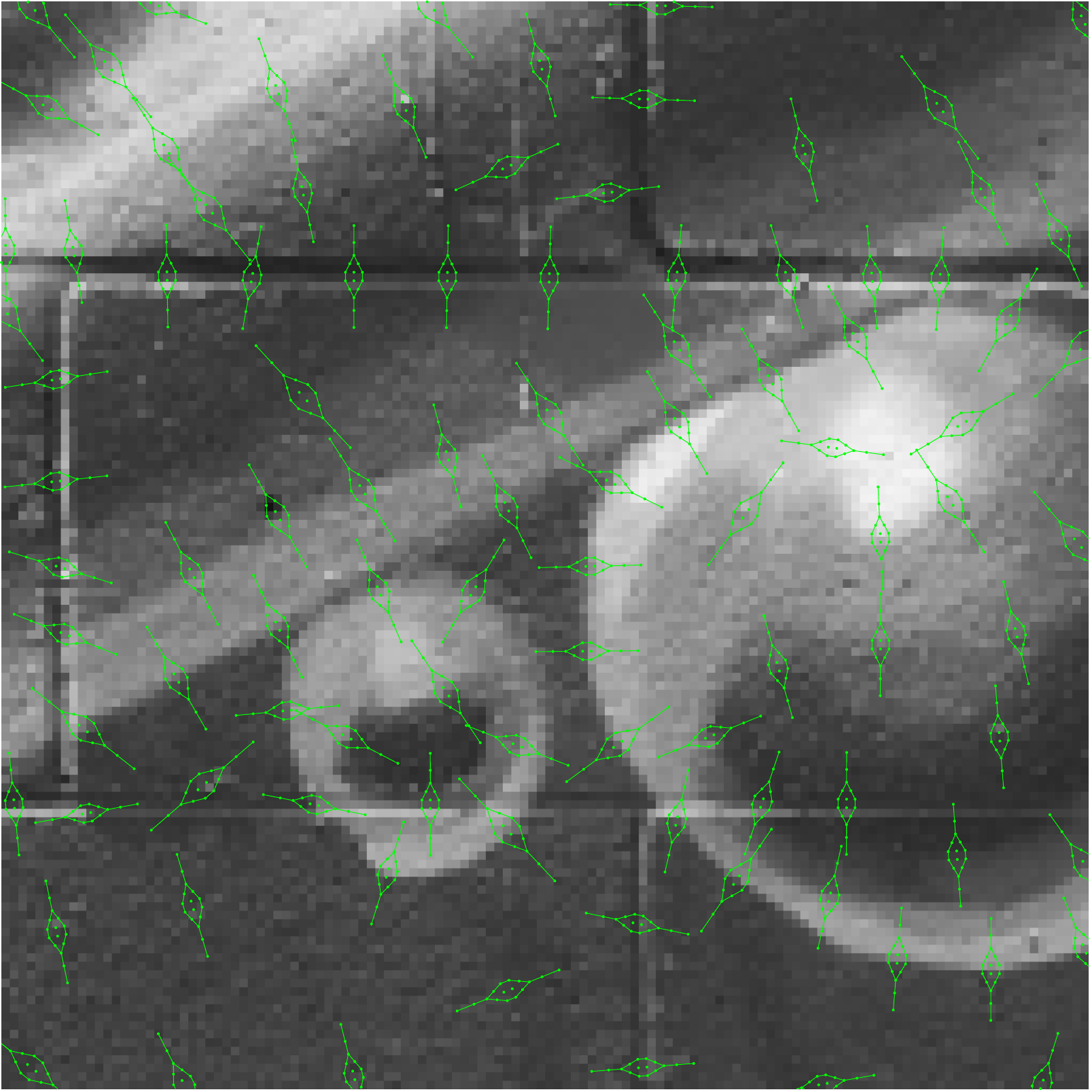}\\
		(a) Input image & (b) Gradient mag. & (c) Edgelet features
	\end{tabularx}
	\caption{{\bf Edgelet extraction.} An input image (a) is first converted to grayscale, from which gradient magnitudes (b) and orientations are computed. Edgelet features (c) are instantiated on the sub-pixel local maxima of gradient magnitude, in the direction of maximum gradient.}
	\label{fig:edgelet_extraction}
\end{figure}

Different patches in a region of constant intensity gradient produce the same NCC cost, therefore a \emph{change} in gradient, \eg an image edge, is required for alignment. The patch selection method used here therefore first computes a list of $P$ features, one per image pixel, the $\ith$ feature consisting of image coordinates, $\coord_i$,  gradients of intensity, $I_i$, in horizontal ($x$) and vertical ($y$) directions, and a score, $\gradscore_i$, where
\begin{align}
\graddir_i &= \begin{bmatrix}\frac{\partial I_i }{\partial x}, \frac{\partial I_i}{\partial y}\end{bmatrix}^\trsp,\\
\gradscore_i &= \log(1+\|\graddir_i\|).
\end{align}
A greedy approach to feature selection is then used to find a well distributed set of $Q$ strong features from that list, by at each iteration selecting the feature that maximizes the product of its score and its squared distance to the closest previously selected feature; see Algorithm \ref{alg:greedy_selection}. A downside of the greedy feature selection is its $\bigO(PQ)$ time complexity; this is mitigated here by only selecting from local maxima of the score, significantly reducing $P$. However, a benefit is that it produces an exact number of features, as well as an ordering, so any smaller number of features can also be generated (by taking the first $R<Q$), at no extra cost. This is used in section \ref{sec:sparse_evaluation} to evaluate the effect of feature density on performance. In practice, a linear time feature selection method can be used, \eg that of Engel \etal~\cite{engel2017direct}. Extraction of the features, here called \emph{edgelets}, is summarized in Figure~ \ref{fig:edgelet_extraction}.

\begin{algorithm}[t]
	\SetAlgoLined
	Given: input features $\{\coord_i,\graddir_i,\gradscore_i\}_{i=1}^P$, output no., $Q$.\\
	Initialize: $d_i := \infty, \forall i$, $\order_1=\argmax_i \gradscore_i$\\
	\For{j~=~2:Q}{
		\For{i~=~1:P}{
			$d_i := \min(d_i, \|\graddir_i-\graddir_{\order_{j-1}}\|_2^2)$\\
		}
		$\order_j := \argmax_i s_i \cdot d_i$\\
	}
	Output: sparse features $\{\coord_{\order_i},\graddir_{\order_i},\gradscore_{\order_i}\}_{i=1}^Q$
	\caption{\label{alg:greedy_selection}Greedy feature selection}
\end{algorithm}

\subsubsection{Patch sample layout}
\label{sec:patch_layout}
Locally normalized patches must consist of at least three samples to be discriminative, since NCC normalization projects out two dimensions from a patch. The number and layout of these samples can have a significant impact on performance. Engel \etal~\cite{engel2017direct} use an 8 sample, diamond layout, with integer pixel coordinates for their sparse SSD cost. The third contribution of this work is to propose an edge-aligned patch layout that improves both convergence rate and time. The novel features of design are:
\begin{itemize*}
	\item 16 sample layout designed to improve both convergence rate and accuracy.
	\item Oriented \wrt to the image edge.
	\item Sub-pixel alignment to the image edge.
	\item Scaled to ensure minimum 1 pixel $\norm\infty$ distance between samples.
\end{itemize*}
The patch sample layout consists of a small cluster of samples around the feature that ensure good texture alignment, and two long, thin arms that enable broad convergence, as shown in Figure \ref{fig:edgelet_extraction}(c). The exact sample locations, relative to an edgelet feature, are
\begin{equation}
\resizebox{0.89\hsize}{!}{$
\offsets = \left[\arraycolsep=2.5pt\begin{array}{cccccccccccccccc} 0 & 0 & 0 & 0.5 & \shortminus0.5 & \shortminus1 &  0 &  1 &   1  &  0  & \shortminus1 & \shortminus0.5 & 0.5 &  0 &  0 & 0\\
6 & 4 & 2.5 & 1.5 & 1.5 & 0.5 & 0.5 & 0.5 & \shortminus0.5 & \shortminus0.5 & \shortminus0.5 & \shortminus1.5 & \shortminus1.5 & \shortminus2.5 & \shortminus4 & \shortminus6\end{array}\right]
$}
\end{equation}
The use of 16 samples suits SIMD vector unit sizes on modern CPUs, allowing for more efficient vectorization of computations. These samples are scaled, rotated and translated per edgelet feature as follows:
\begin{equation}
\offsets_i = \frac1{\max\left(|\frac{\partial I_i }{\partial x}|, |\frac{\partial I_i}{\partial y}|\right)} \begin{bmatrix} -\frac{\partial I_i}{\partial y} & \frac{\partial I_i}{\partial x} \\ \frac{\partial I_i}{\partial x} & \frac{\partial I_i}{\partial y} \end{bmatrix}\offsets + \coord_i \ones^\trsp.
\end{equation}
The scaling ensures a minimum 1 pixel distance between samples in any direction, while the rotation ensures that the patch layout is aligned perpendicular to the image edge it straddles. In addition, the feature location, $\coord_i$, is refined perpendicular to edge direction, by fitting a quadratic function to the intensity gradient magnitude across the edge, and shifting the feature to the maximum of that quadratic. Patch samples are therefore not generally at integer locations (or rather pixel corners) in the image.

\begin{figure*}[t]
	\setlength{\imwidth}{0.195\linewidth}
	\subfloat[][One image each from 6 of the 7 sets in Graffiti2.]{
		\begin{tabularx}{0.595\linewidth}{@{}*{2}{c@{\extracolsep{\fill}}}c@{}}
			\includegraphics[width=\imwidth]{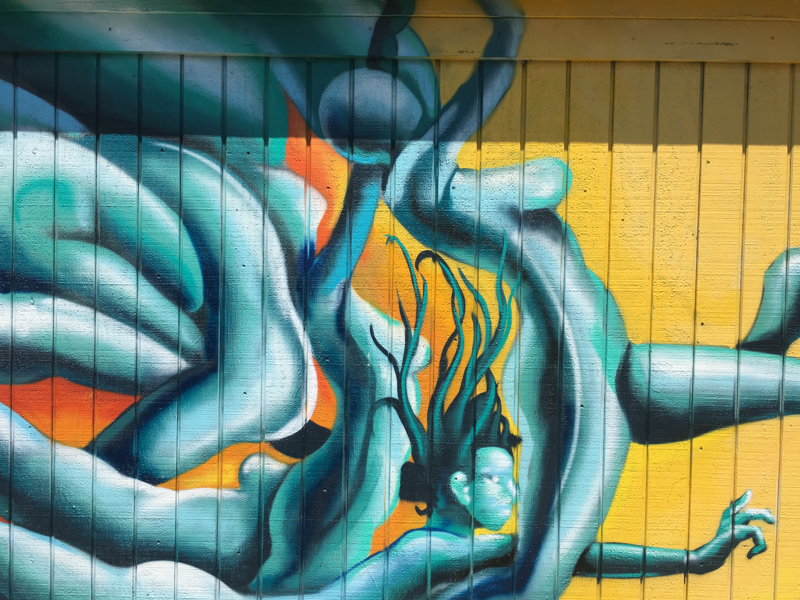} & \includegraphics[width=\imwidth]{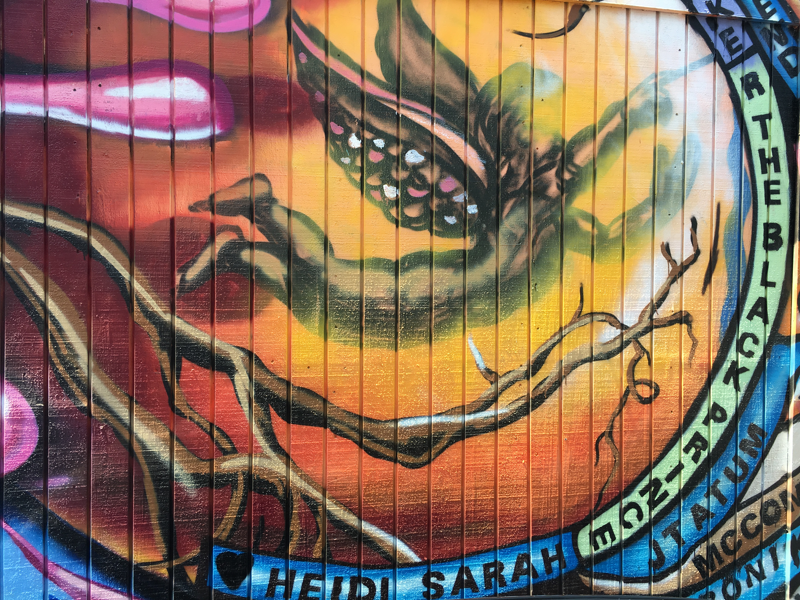}  & \includegraphics[width=\imwidth]{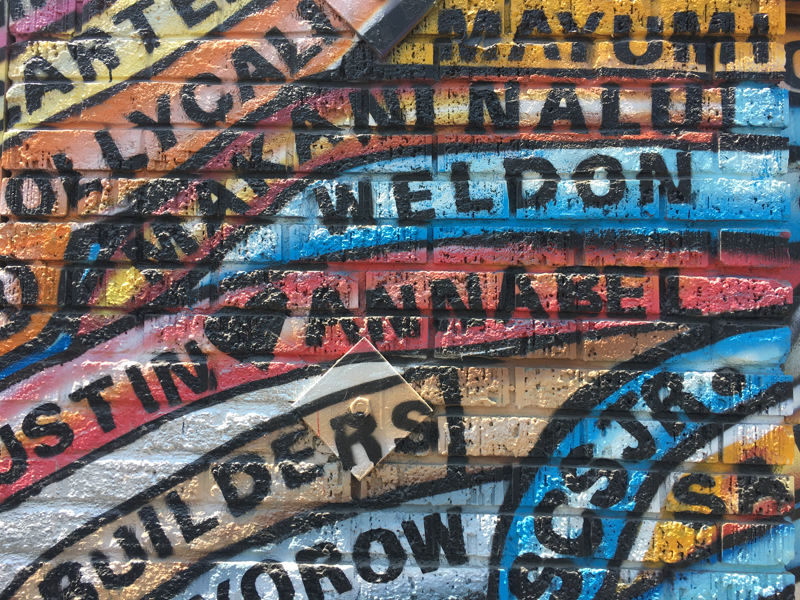} \\
			\includegraphics[width=\imwidth]{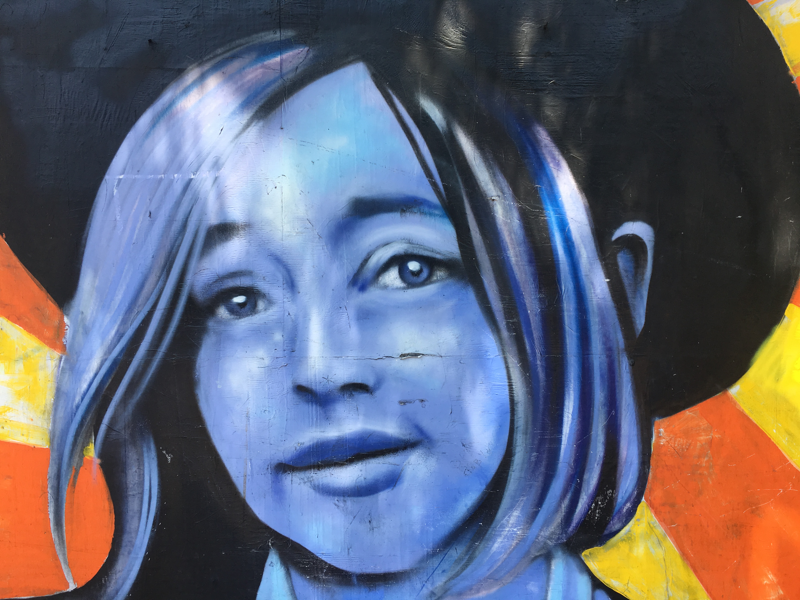} & \includegraphics[width=\imwidth]{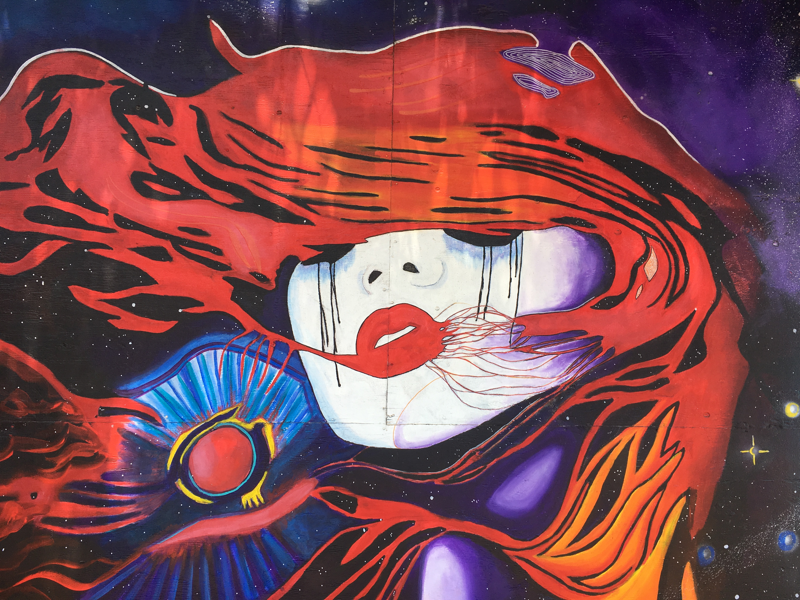}  & \includegraphics[width=\imwidth]{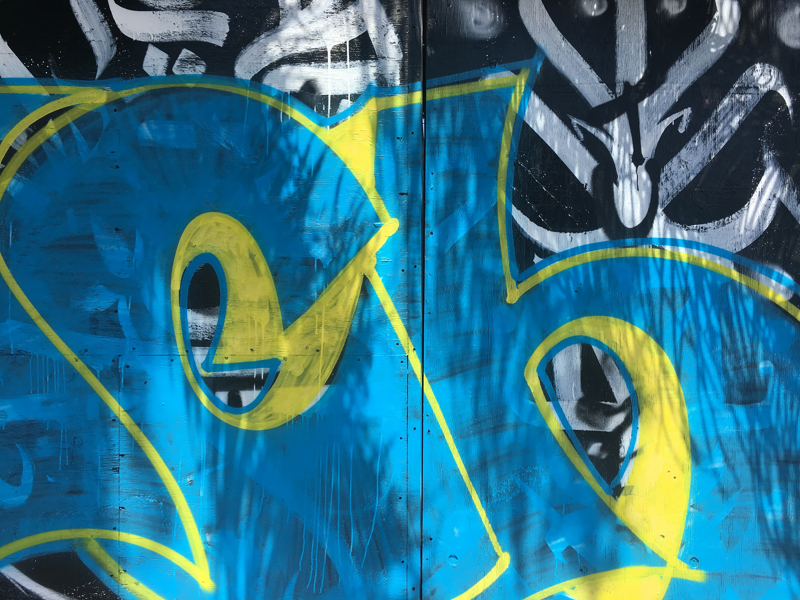}
	\end{tabularx}}
	\hspace{0.01\linewidth}
	\subfloat[][All 4 images from the remaining set of Graffiti2.]{
		\begin{tabularx}{0.398\linewidth}{@{}*{2}{c@{\extracolsep{\fill}}}c@{}}
			\includegraphics[width=\imwidth]{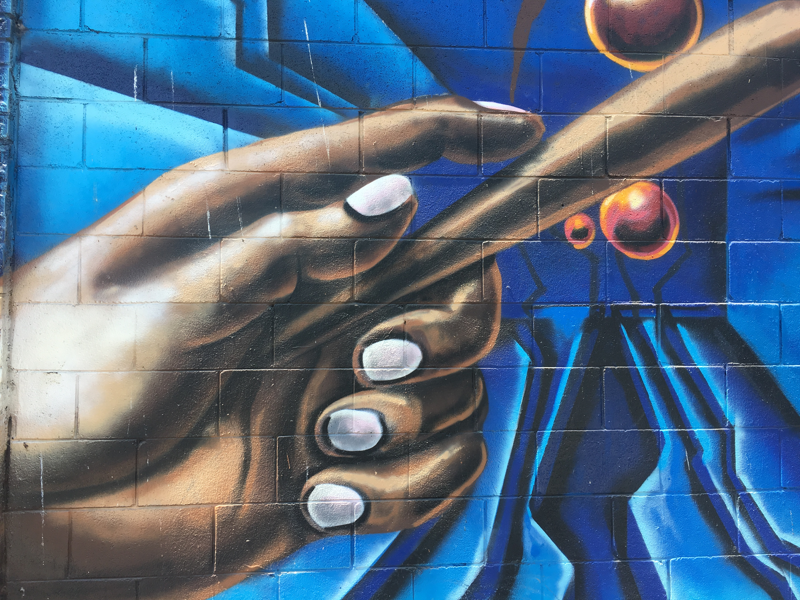} & \includegraphics[width=\imwidth]{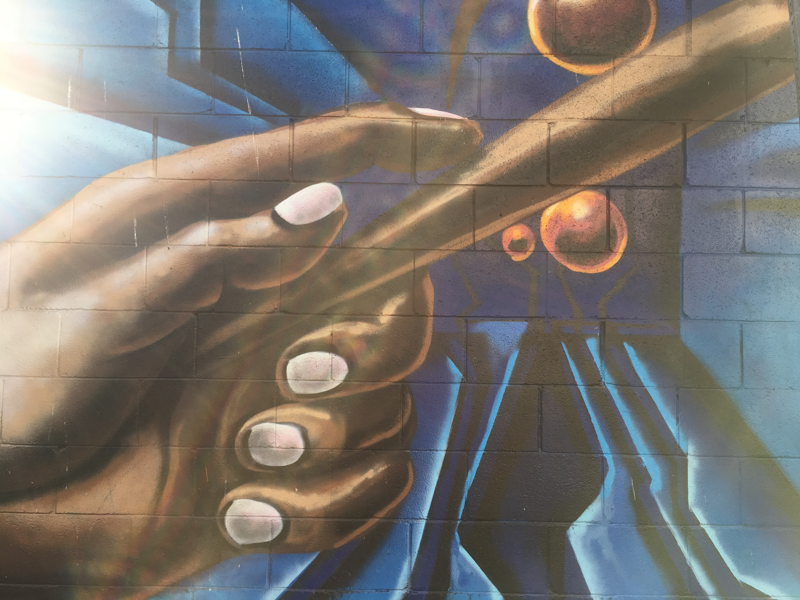}\\
			\includegraphics[width=\imwidth]{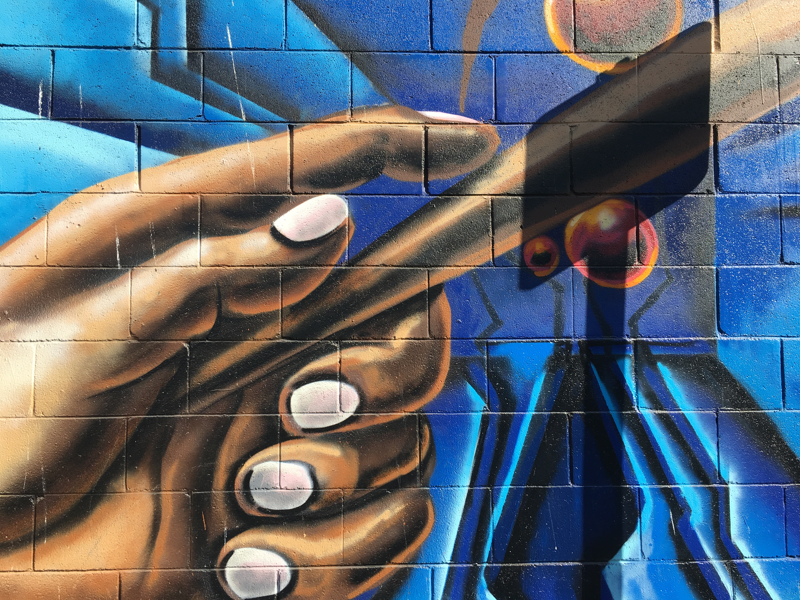} & \includegraphics[width=\imwidth]{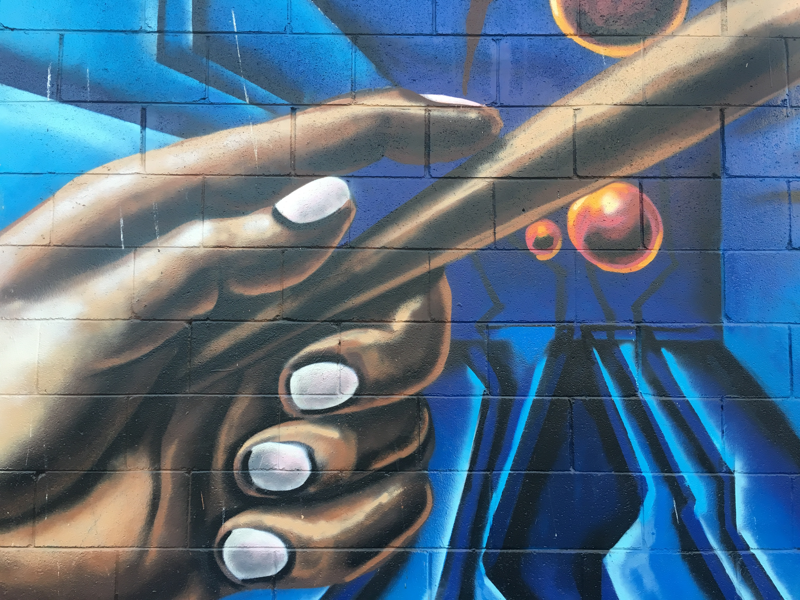}
	\end{tabularx}}
	\caption{{\bf Graffiti2 dataset.} A dataset of 7 image sets (a), of 4 600$\times$800$\times$3 images each (b), of approximately planar regions of outdoor graffiti, taken from roughly the same position and under different lighting conditions due to capture at different times of the day, is used and made available~\cite{woodforddatasets}. All images were captured with an iPhone 6s (with negligible lens distortion) as 12Mpixel JPEG photos, then downsampled and saved losslessly.}
	\label{fig:dataset}
\end{figure*}

\subsection{Implementational details}

\subsubsection{Solving the linear system}
\label{sec:solve_linear}

Eq.~(\ref{eqn:linear_system}) is commonly solved using the normal equations:
\begin{equation}
	\label{eqn:normal_equations}
	\deltawarp = -\underbrace{\left(\jac^\trsp\jac\right)^{-1}\jac^\trsp}_{\textrm{Pseudo-inverse,~}  \jac^+}\error.
\end{equation}
In this case, it is worth considering whether the problem is well conditioned. Each normalized patch provides $\patchlen-2$ constraints, since the mean and variance normalizations both remove a constraint, therefore if $\numpatches(\patchlen-2)\ge\warpdims$, a unique solution to equation~(\ref{eqn:linear_system}) will generally exist. However, if image texture is lacking, or parallel to epipolar lines, the problem can be poorly conditioned. To regularize the problem, the minimum norm solution to equation~(\ref{eqn:linear_system}) is solved. This can be done by computing the Moore-Penrose pseudo inverse~\cite{moore1920pseudoinverse}, $\jac^+$, in equation (\ref{eqn:normal_equations}). However, this approach is relatively slow, therefore only used here when the Jacobian is fixed (\ie when using inverse compositional Jacobians and no robustifier). In other cases, an orthogonal decomposition is used to directly compute the minimum norm solution.\!\!\footnote{The \texttt{lsqminnorm} function in \matlab is used, as follows: $\deltawarp = -\texttt{lsqminnorm}(\jac, \error, 10^{-8})$.}
Alternatively, the Levenberg-Marquardt algorithm~\cite{tingleff2004methods}, which also regularizes the linear problem, could be used.

\subsubsection{Convergence criteria}
\label{sec:convergence_criteria}
In the implementation used, the update of equation (\ref{eqn:linear_system}) is repeated until either $\|\deltawarp\|_\infty < 10^{-6}$,  the least squares cost fails to go below the minimum found for three consecutive iterations, the reduction of the minimum is in the range [0,0.01]\%, or 100 iterations elapse.

\subsubsection{Handling homogeneous regions}
\label{sec:homogeneous_regions}
When a homogeneous patch is passed through $\patchtransform$, it will result in the vector $\zeros/0$, which is undefined; furthermore, $\jacncc$ (\eqnref{jac_ncc}) will be infinite. Using the global cost, this would be a rare occurrence, since patches used for alignment are generally large, and would be expected to contain at least some texture. However, using the local cost, patches used can be of arbitrary length (down to three samples) and cover small regions of an image, therefore it is quite possible that some of these patches will enter homogeneous regions in the course of an optimization. The solution used here is that if $\SD=0$, it is set to 1 instead, such that the output patch and Jacobian are both $\zeros$. As a result, the (unrobustified) cost of a source patch in a homogeneous region, assuming the target patch is not homogeneous,\!\!\footnote{It is redundant to have a target patch in a homogeneous region, since this patch offers no gradient information.} will always be 1. For this reason, in order to avoid homogeneous regions creating local minima, it is advisable to use a robustification function which truncates the least squares cost at or below 1.

\section{Evaluation}
\label{sec:evaluation}
Homography-based image alignment is used to evaluate the photometric error and optimization framework presented herein. Homographies model planar motion captured with a pinhole camera, and as such is the simplest geometrically accurate motion model under perspective projection. Planar targets are therefore used in the evaluation. 
Furthermore, the reasonably high dimensionality of this problem's state space (\dims{8}) indicates performance on the kind of problems for which photometric least squares methods remain popular, such as active appearance models~\cite{matthews2004active} and direct odometry~\cite{engel2017direct}. Experiments validate the value of the contributions of this work, using a statistical evaluation of performance similar to previous works~\cite{baker2004lucas,evangelidis2008parametric}. Further qualitative results on multi-resolution tracking in videos demonstrates the real-world applicability of this method. All experiments and results are completely reproducible, via publicly available source code and datasets.\!\!\footnote{\matlab source code (which also downloads datasets) is available at: \url{https://github.com/ojwoodford/image-align}.}

\subsection{Single resolution, quantitative evaluation}
\label{sec:quantitative_eval}
A quantitative evaluation is used to show the relative benefits of the approach proposed here versus other NCC optimization frameworks, as well as other photometric errors.

\subsubsection{Dataset \& metrics}
To ensure that results are broadly applicable, experiments should be run on a range of textures from real photos, with locally varying intensities across images of the same scene. A new dataset of 7 sets of 4 images, shown in Figure \ref{fig:dataset}, with real-life changes in lighting due to capture at different times of day, is presented. Ground truth homographies between each pair of images within each set are computed, using a standard feature matching with RANSAC approach~\cite{brown2007autostitch}, followed by photometric alignment using the Census Bitplanes cost~\cite{alismail2016direct}.

Metrics are computed over a set of alignments of 48$\times$48 pixel image regions,\!\!\footnote{48 is a multiple of 2, 3, 4, 6 \& 8, used in sec.~\ref{sec:block_size_evaluation}.} generated as follows: 100 regions, that each contain at least 100 edgelets, and are fully visible in the 3 other images of the same scene, are randomly selected from each image. A perturbation is generated for each region, by applying a random shift, drawn from a normal distribution, to each corner separately, then scaling the perturbation so that the mean shift per corner is 0,..,10 pixels, creating starting distances of increasing magnitude for each region. 
Each perturbation of each region is warped into each other image in the set, using the ground truth homography, from which the initial warp from target to source is then computed~\cite[alg. 4.2]{hartley2004multiviewgeometry}. This creates a total of 100$\times$11$\times$4$\times$3$\times$7 = 92,400 alignment test cases between different images.

A number of metrics are used to evaluate the relative performance of methods:
\paragraph{Error} Each homography, both ground truth and estimated, transforms the target region corners into source image coordinates, using equation (\ref{eqn:warp}). Error is the maximum of the four Euclidean distances (in pixels) between the ground truth region corners and the computed region corners in the source images.
\paragraph{Convergence \emph{or} recall} These two terms refer to the percentage of tests that have an error below a given threshold, 1 pixel for convergence, and a variable threshold in recall \vs error plots (computed at an initial mean corner error of 4 pixels).
\paragraph{Mean convergence time \& iterations to converge} The computation time and number of iterations of each test is recorded. All tested methods are implemented in the same \matlab framework, such that timings are a good indicator of relative, if not absolute, performance. Statistics for those tests that meet the 1 pixel error convergence criterion are averaged, to produce these two measures.
\paragraph{Iteration time} This is computed as the total computation time over all tests, divided by the total number of iterations over all tests.

All three composition methods described in section \ref{sec:jacobian}, denoted FWD, INV and ESM for forward compositional, inverse compositional and efficient second-order minimization methods respectively, are tested.

\subsubsection{NCC optimizer comparison}
\label{sec:optim_comp}
As discussed in section~\ref{sec:previous_ncc}, many optimizers of NCC have been proposed. Experiments compare the following most recent and effective methods:
\begin{itemize*}
	\item \textbf{LSNCC}: The least squares NCC optimizer proposed here, which is also analytically equivalent to, though implementationally simpler than, that of Scandaroli \etal~\cite{scandaroli2012ncc} (except for ESM; see Appendix~\ref{sec:scandaroli}).
	\item \textbf{SMR}: The ESM alternative, combining forward and inverse Jacobians, proposed by Scandaroli \etal~\cite{scandaroli2012ncc}.\!\!\footnote{SMR references the authors Scandaroli, Meilland \& Richa~\cite{scandaroli2012ncc}.}
	\item \textbf{ZNSSD}: ``Zero-mean, normalized, sum of squared differences'', proposed by Pan \etal~\cite{pan2013znssd}, which approximates the Jacobian.
	\item \textbf{ECCM}: ``Enhanced correlation coefficient maximization'', proposed by Evangelidis \& Psarakis~\cite{evangelidis2008parametric}.
	\item \textbf{GB}: A gain \& bias intensity model optimizer~\cite{li2018photometriclie}, very similar in effect to the normalization of NCC.
\end{itemize*}
Experiments cover three costs: the global cost of eq.~(\ref{eqn:global_cost}) and the local cost of eq.~(\ref{eqn:local_cost}), first without robustification (\ie $\robustify(s) = s$), then with robustification or residual weighting of various kinds. The GB intensity parameters are also applied globally or locally, as appropriate. These experiments all involve sampling a dense grid at pixel corners, 
as opposed to sparse edgelets, which are evaluated later. 

\begin{figure*}
	\setlength{\imwidth}{0.327\linewidth}
	\begin{tabularx}{\linewidth}{@{}*{2}{c@{\extracolsep{\fill}}}c@{}}
		\subfloat[][Globally normalized cost.]{\includegraphics[width=\imwidth]{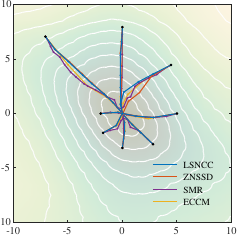}} &
		\subfloat[][Locally normalized cost.]{\includegraphics[width=\imwidth]{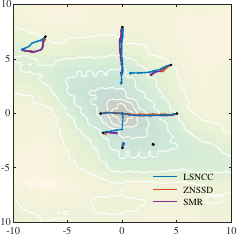}} &
		\subfloat[][Robustified, locally normalized cost.]{\includegraphics[width=\imwidth]{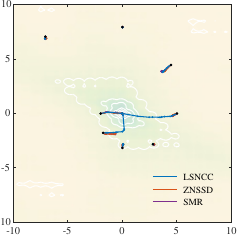}}
	\end{tabularx}
	\caption{{\bf Surfaces of NCC cost functions.} NCC cost surfaces (contour map) for three  of the costs tested, computed on an 48$\times$48 pixel patch shifted over a 20$\times$20 pixel region centered on itself (the patch from fig. \ref{fig:edgelet_extraction}(a) was used). Each plot additionally shows the trajectory of \dims{2} translation optimizations for the ESM variant of each optimizer evaluated, starting from 8 different points (black dots).}
	\label{fig:cost_surface}
\end{figure*}

Figure~\ref{fig:cost_surface} qualitatively compares these costs through a visualization of their cost surfaces. The global cost (a) has a single, large convergence basin, while the unrobust local cost (b) has a smaller basin surrounded by a plateau with some local minima. The robust cost (c) has the smallest basin and a more homogeneous plateau. Overlaid optimzer trajectories show that the methods take similar, though clearly different, paths to the optimum, and that LSNCC reached the optimum more often in the local and robustified cases. Quantitative results for each case are now discussed.

\paragraph{Global cost}
Figure \ref{fig:all_quantitative}(a) shows results for globally normalized optimizations. The first thing to note is that the convergence rate at an initial corner error of zero pixels is surprisingly low, at around 60\%, but that it decreases slowly after that, with 40\% convergence at a start distance of 10 pixels. This indicates that global normalization often cannot handle real-world lighting changes, which are local in nature, causing correct alignments to actually diverge. However, it has broad convergence properties. It should also be noted that methods with the lowest convergence rate at large starting distances (here, the GB method with INV composition) tend to have a higher convergence rate at zero starting distance, since the narrower convergence basin is less likely to cause the alignment to diverge to another minimum, even a better one. For this reason, convergence rates at zero should not be used as a measure of performance, and are provided simply for completeness.

In terms of convergence rates, LSNCC very marginally outperforms other methods for ESM and INV compositions, though performs worse with the FWD composition. However, this latter composition should be avoided if possible anyway, since it offers the lowest convergence rates along with the slowest convergence times. Interestingly, the GB and ECCM methods perform noticeably worse than LSNCC and ZNSSD for the INV composition at larger starting distances. In terms of computation time, LSNCC and ZNSSD are equally fast using INV composition, but LSNCC converges more slowly using ESM, due to the extra computation required to account for normalization in the Jacobian (precomputed for INV). ECCM requires a more expensive adjustment, slowing ESM slightly more still, but this adjustment is also required in the INV case, slowing convergence down significantly. The GB method introduces a further two variables to optimize over, making it the slowest method for ESM, though only marginally slower than LSNCC \& ZNSSD for INV. But in general there is not much to choose between the methods here, especially LSNCC and ZNSSD.

\paragraph{Local cost}
Figure \ref{fig:all_quantitative}(b) shows results for locally normalized, but unrobustified, optimizations. As discussed in section \ref{sec:previous_ncc}, the ECCM method does not extend to this scenario, therefore is omitted. The first thing to note is that convergence rates at zero starting distance increase to almost 80\%, from under 60\% for the global cost. This demonstrates that local normalization better models the lighting variations of real-world scenes. The second thing to note is that there is a much broader spread in performance, with LSNCC performing significantly better on convergence and recall than both ZNSSD and GB, for all compositions. The improvement over ZNSSD highlights the need to accurately compute the Jacobian in the locally normalized case. Per iteration computation times are similar to the global cost for both LSNCC and ZNSSD, while GB is substantially slower, due to the large increase in variables caused by normalizing (or rather modelling gain \& bias) locally. This increase also suggests a reason for its dramatic degradation in performance; a larger parameter space leads to more local minima to get trapped in. Optimization time for LSNCC in the ESM case is about the same as ZNSSD, since it converges in slightly fewer iterations, to offset the slower per iteration time. The fact that GB appears to become relatively faster at larger starting distances is likely due to the fact that only successful optimizations contribute to the time, which induces a selection bias; methods with smaller minima will tend to converge only on tests that converge more easily, and therefore also faster.

\begin{figure}
	\setlength{\imwidth}{0.32\linewidth}
	\centering
	\begin{tabularx}{\linewidth}{@{}*{2}{c@{\extracolsep{\fill}}}c@{}}
		\includegraphics[width=\imwidth]{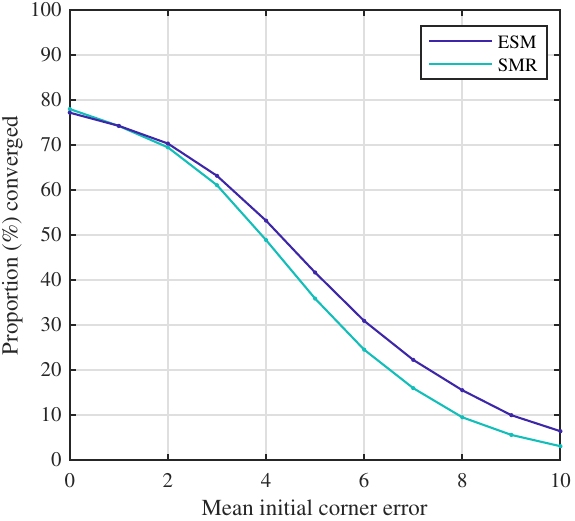} & \includegraphics[width=\imwidth]{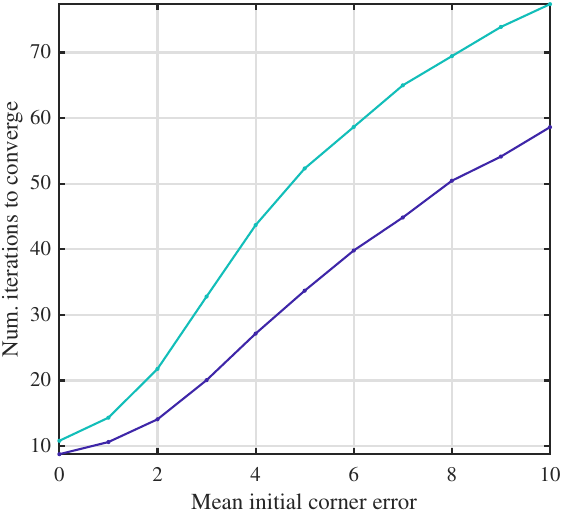} & 
		\includegraphics[width=\imwidth]{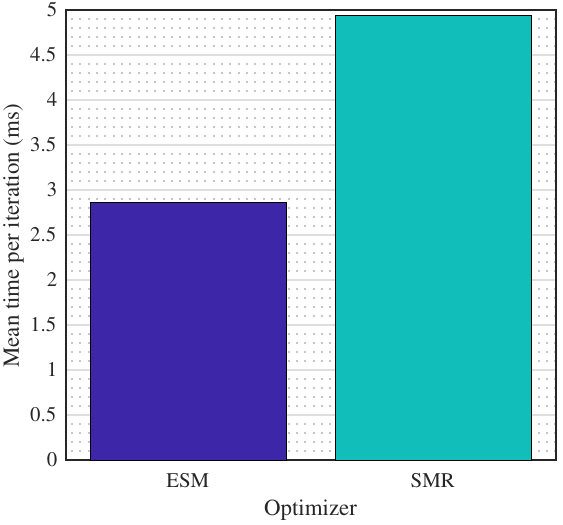}\\
		(a) Convg.\ rate & (b) Iters.\ to convg. & (c) Iteration time 
	\end{tabularx}
	\caption{{\bf ESM \vs Scandaroli \etal.} Performance, in terms of (a) convergence rates, (b) iterations to converge and (c) iteration time, for ESM~\cite{benhimane2004esm} and the combined update of Scandaroli \etal (SMR)~\cite{scandaroli2012ncc}, using an unrobust, local cost. 
		\label{fig:scandaroli_esm}}
\end{figure}

The SMR hybrid method~\cite{scandaroli2012ncc}, which differs mathematically from ESM, as shown in Appendix \ref{sec:scandaroli}, was also tested on this cost. Their update performs marginally worse than LSNCC ESM in terms of convergence rates, as shown in Figure \ref{fig:scandaroli_esm}(a), but requires significantly more iterations (b), and more time per iteration also (c), since the least squares formulation can be solved without constructing the normal equations, unlike SMR.

\paragraph{Robust, local cost}
This work proposes the use of the Geman-McClure kernel to robustify the cost of each patch. Results of optimizations using this kernel are shown in Figure \ref{fig:all_quantitative}(c), for both the LSNCC and ZNSSD optimization frameworks. Robustification can be seen to offer two benefits over the unrobustified, locally normalized cost: convergence rates at zero starting distance improve slightly, to over 80\%, but more importantly, this convergence basin is broadened, with LSNCC ESM achieving over 70\% convergence at a starting distance of 4 pixels, compared to under 55\% for the unrobustified cost. The improvement in convergence and recall rates of LSNCC over ZNSSD mirror those of the unrobustified cost, re-iterating the need to accurately compute the Jacobian.

Robustification effectively introduces a weighting per normalized patch, based on its error.  The cost is therefore also compared with previously proposed weighting schemes. Irani \& Anandan~\cite{irani1998robust} weight each patch using the determinant of its Hessian, which in this work is approximated by $\jac^\trsp \jac $. Figure \ref{fig:all_quantitative}(c) shows that this scheme, denoted Det.~H, performs extremely poorly in terms of convergence rates, when applied to the LSNCC cost. This might be due to using Gauss' approximation of the Hessian, but might also be due to the fact that the \dims{8} problem tackled here is a much higher dimensionality than the \dims{2} alignment problem used in the original work~\cite{irani1998robust}. It should also be noted that this scheme makes no sense in the INV case, since the Hessian doesn't change. Another scheme~\cite{scandaroli2012ncc}, denoted SMR w., combines the Huber kernel with a k-means clustering of NCC scores. This performs noticeably worse than both LSNCC and ZNSSD, and is also slower to compute. These two results highlight the need for robustification grounded in theory, as opposed to ad-hoc schemes.

\subsubsection{Photometric error comparison}
\label{sec:error_comp}
The previous section investigated the relative performance of different optimizers on the same photometric costs. This section investigates performance of different photometric errors. LSNCC here refers to the robust, locally normalized variant, unless otherwise stated.

\begin{figure*}
\setlength{\imwidth}{0.248\linewidth}
\centering
\begin{tabularx}{\linewidth}{@{}*{3}{c@{\extracolsep{\fill}}}c@{}}
	\includegraphics[width=\imwidth]{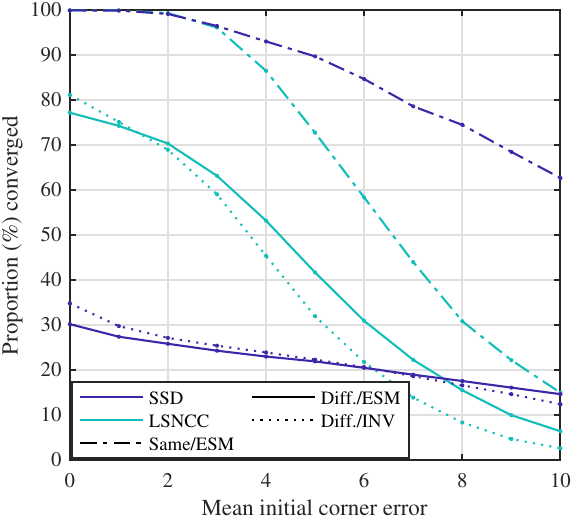} & \includegraphics[width=\imwidth]{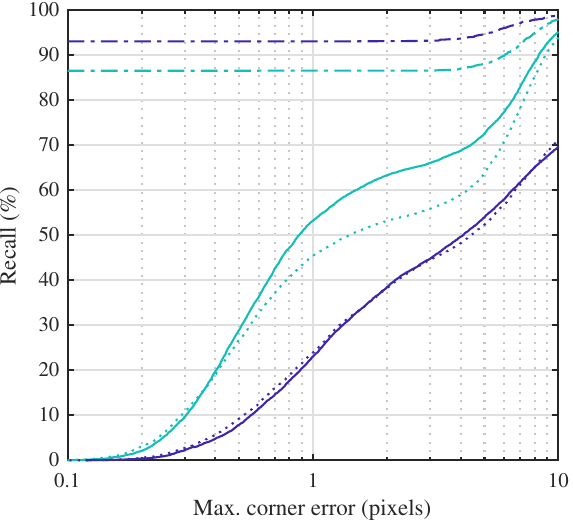} & \includegraphics[width=\imwidth]{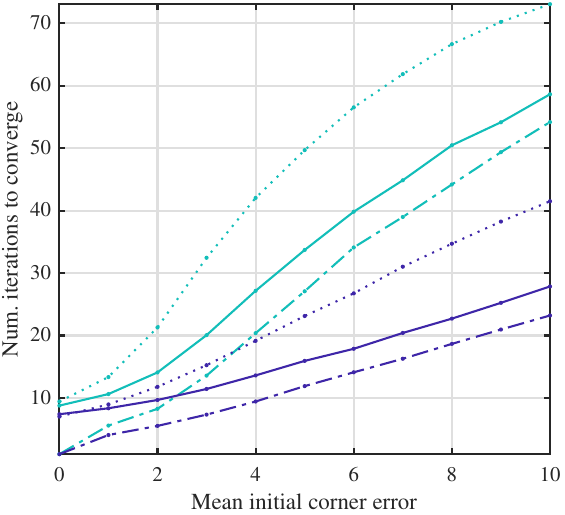} & \includegraphics[width=\imwidth]{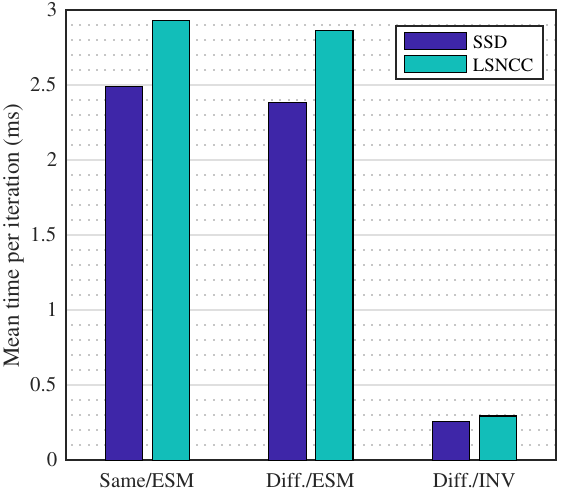} \\
	(a) Convergence rates & (b) Recall \vs error & (c) Mean iterations to converge & (d) Mean iteration time
\end{tabularx}
\caption{{\bf Comparison to SSD.} The locally normalized NCC cost with 6$\times$6 blocks is compared to the SSD cost, for two cases: where the source and target images are identical (\ie the same image), or different. Neither cost uses any robustifier. The ESM update is used.
	\label{fig:ssd_quantitative}}
\end{figure*}

\paragraph{Versus SSD}
The most commonly used photometric cost is the sum of squared differences (SSD). It assumes constant brightness, which holds true if there is no change in lighting and capture settings between images, and the scene is perfectly Lambertian. Often this will be a reasonable assumption, for example when tracking in video over a short period of time. For this reason it is instructive to evaluate how much is lost by using  LSNCC in such a situation. 
In terms of convergence, Figure \ref{fig:ssd_quantitative}(a) shows that in the case that source and target are the same image (dash-dot lines), both costs provide perfect convergence at the smallest starting distances, but that the convergence rate of LSNCC drops off significantly faster with starting distance. This latter trend extends to the standard case (solid and dotted lines), where source and target are different images (with locally varying changes in intensity), but here LSNCC now has a much higher convergence rate than SSD at smaller starting distances, as expected.
In terms of computational efficiency, it can be seen (Fig.\ \ref{fig:ssd_quantitative}(d)) that each iteration of LSNCC is slower than SSD, by 14\% in the INV case, and 18\% for ESM. However, a more significant slowdown is caused by the greater number of iterations that LSNCC requires to converge (Fig.\ \ref{fig:ssd_quantitative}(c)), about twice as many as for SSD starting from a 4 pixel corner error.
Therefore LSNCC trades off robustness to intensity variations against both convergence basin size and also computation time.

\paragraph{Versus transformation-based methods}
We saw in the previous section that LSNCC is the best optimizer of the robust, locally normalized NCC cost. However, it remains to be seen whether robust, locally normalized NCC is the best photometric error for handling local intensity variations. Here this cost is compared against two other locally lighting invariant photometric costs: Descriptor Fields~\cite{crivellaro2014descriptorfields} (first order version) and the Census Bitplanes transform~\cite{alismail2016robust}. Both are examples of image transformation approaches, and showed state of the art results when published. While there are more recent, deep learned image transformations~\cite{chang2017clkn,vonstumberg2020gn}, neither code nor weights for these methods are publicly available,\!\!\footnote{Personal correspondence with Che-Han Chang~\cite{chang2017clkn}, \fst{21} August 2018, and Lukas von Stumberg~\cite{vonstumberg2020gn}, \nth8 May 2019.} and they are also specific to the data they are trained on, so are not evaluated here. 

All the methods are implemented and evaluated within the same framework presented here (\ie same sample locations, warp parametrization, least squares solver, convergence criteria \etc), using ESM.
Although Census Bitplanes as published applied the photometric transformation after warping~\cite{alismail2016robust}, the transformation is non-differentiable, therefore here the transformation is applied prior to warping, as this allows forwards compositional derivatives to be computed (enabling ESM), and also only requires the transformation to be computed on the target image once, rather than every iteration.

The Census Bitplanes (CB) cost exhibits high convergence rates at small starting distances, but this performance drops off rapidly, implying the cost has a narrow basin of convergence; this can be mitigated using a coarse-to-fine strategy (used in sec.\ \ref{sec:tracking_videos}). Conversely, Descriptor Fields (DF) has a lower convergence rate at smaller starting distances, but this rate falls off more slowly at larger distances, indicating lower accuracy but greater robustness. However, dense (6$\times$6) LSNCC (which uses the same sample points) has a higher convergence rate than both CB and DF over all starting distances $>$0 tested, for any given composition, providing the best accuracy \emph{and} robustness. In addition, dense LSNCC converges faster than both CB and DF over all starting distances and compositions, except for INV DF. The robustification of LSNCC leads it to have a slower per iteration computation time in the INV case, since the pseudo-inverse cannot be precomputed, whereas it can for both CB and DF costs. However, for other compositions, the lower number of channels provides LSNCC with a significant speed advantage.

\paragraph{Grid block size}
\label{sec:block_size_evaluation}
The dense (regular grid-based samples) LSNCC cost normalizes 6$\times$6 blocks by default. Figure \ref{fig:esm_quantitative}(a) presents results when setting block lengths to 2, 3, 4, 6 and 8 samples. It can be seen that larger normalization regions lead to broader, and faster, convergence, but a lower rate of convergence at zero starting distance. This demonstrates a trade-off between accuracy (with smaller regions) and robustness (with larger regions). With per iteration time impacted negligibly by block size (since the total number of samples remains constant), the faster convergence times of larger block sizes is down to them converging in fewer iterations.

\paragraph{Sparse features}
\label{sec:sparse_evaluation}
Section \ref{sec:feature_selection} introduced a sparse selection of oriented, 16-sample patches, that are used instead of the dense grid of sample patches. 100 sparse features are extracted from each 48$\times$48 test region (a 31\% reduction in samples).

Figure \ref{fig:all_quantitative}(d) compares this sparse LSNCC cost with the dense cost. For INV and ESM compositions, the former has a slightly lower convergence rate at small starting distances, but the rate drops off slower with increasing distance (col.\ 1), suggesting an increase in convergence basin size. However,  with FWD composition, convergence rates with sparse features are better across all starting distances (col.\ 1) and corner error thresholds (col.\ 2). One possible explanation for this is that the sparse samples are not placed on pixel corners, where the range of the linear approximation provided by INV gradients is maximized. Given the modest reduction in number of samples here, sparse features recorded only a modest reduction on convergence and iteration times (cols.\ 3 \& 4). This reduction can be more significant for larger regions; see sec.\ \ref{sec:tracking_videos}.

Figure \ref{fig:esm_quantitative}(b) compares results using different numbers of sparse features (for ESM only), as well as comparing against the 8-pixel sample layout proposed in Direct Sparse Odometry (DSO)~\cite{engel2017direct} (using the same feature locations). A trade-off between accuracy (with more features; col.\ 1) and convergence speed (with fewer features; col.\ 3) is clearly visible, and entirely expected. Also worth noting is that the DSO sample layout has a similar convergence rate at small starting distances, but that rate drops off faster as distance increases (col.\ 1). In addition, despite using half as many samples, DSO's layout offers only marginally faster time to convergence (col.\ 3) than the layout proposed here.  These results suggest that while accuracy is determined by the number of features, convergence basin size is determined by sample layout.

\subsubsection{Handling occlusions}
\label{sec:occlusions}
\begin{figure*}
	\setlength{\imwidth}{0.24\linewidth}
	\centering
	\begin{tabularx}{\linewidth}{@{}p{10pt}@{\hspace{2pt}}*{3}{c@{\extracolsep{\fill}}}c@{}}
		\rotatebox{90}{\footnotesize \hspace{3pt}(a) Globally normalized optim.} & \includegraphics[width=\imwidth]{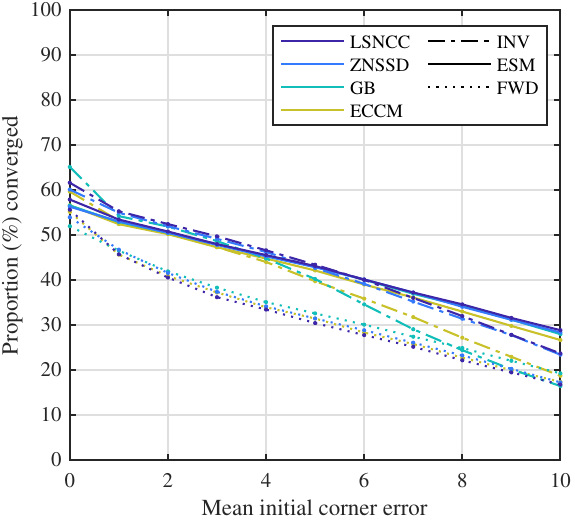} & \includegraphics[width=\imwidth]{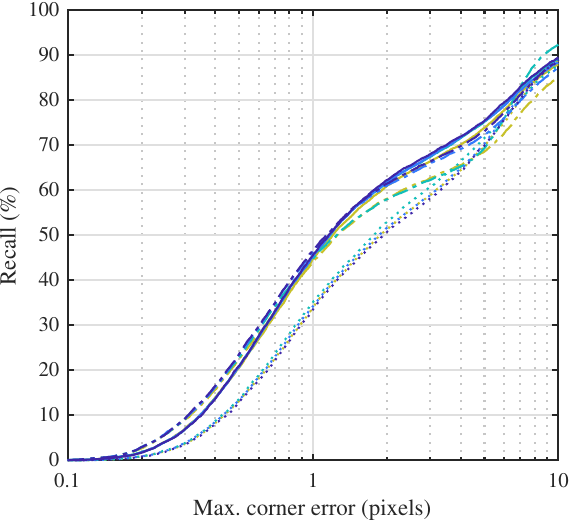} & \includegraphics[width=\imwidth]{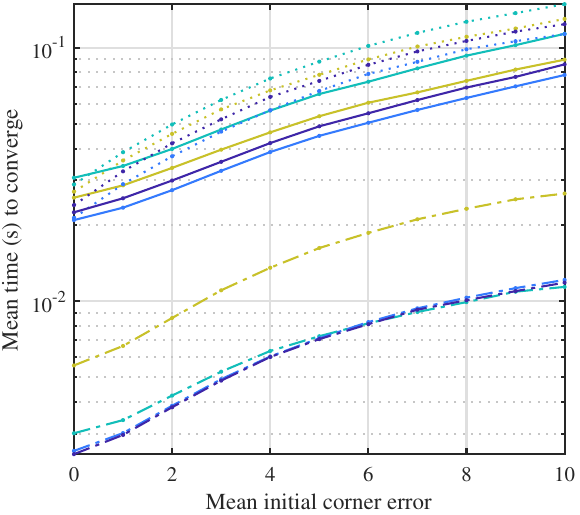} & \includegraphics[width=\imwidth]{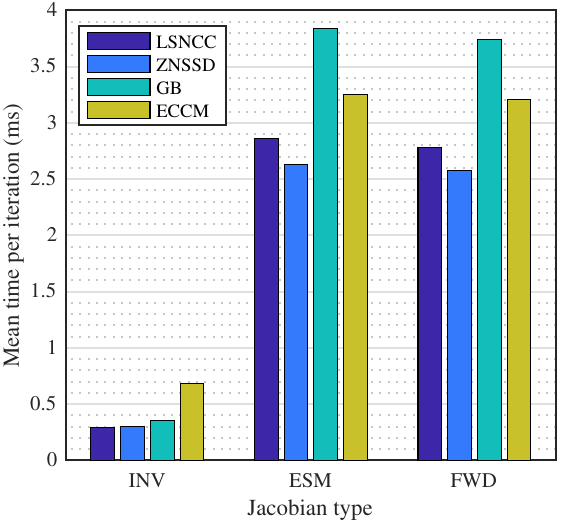} \\
		\rotatebox{90}{\footnotesize \hspace{7pt}(b) Locally normalized optim.} & \includegraphics[width=\imwidth]{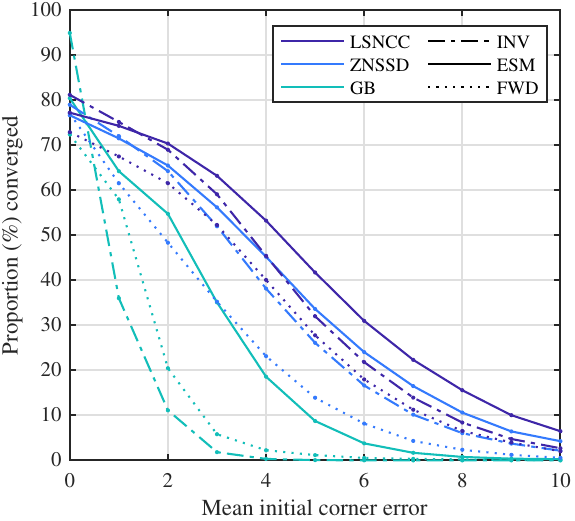} & \includegraphics[width=\imwidth]{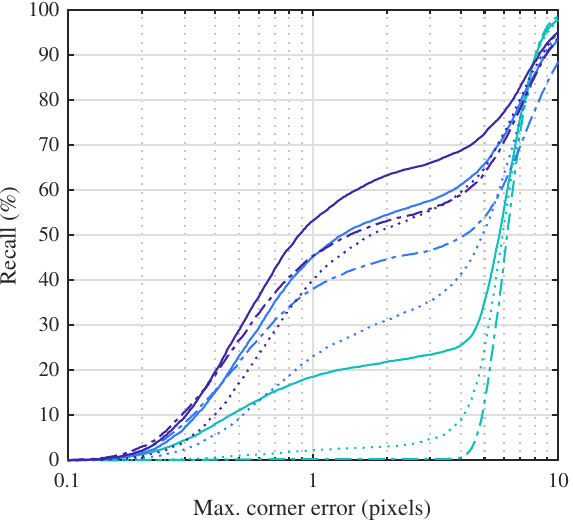} & \includegraphics[width=\imwidth]{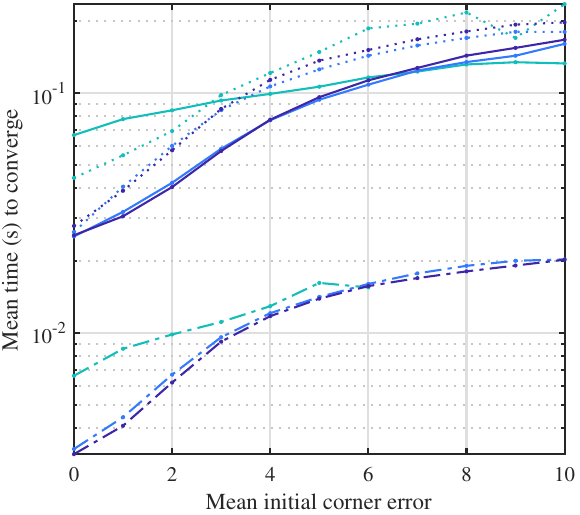} & \includegraphics[width=\imwidth]{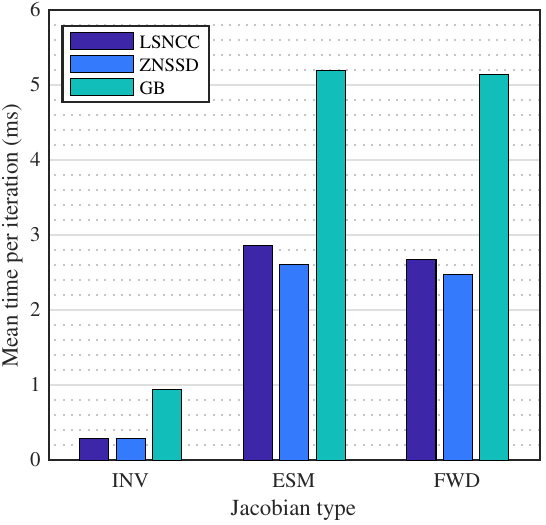} \\
		\rotatebox{90}{\footnotesize \hspace{12pt}(c) Robust/weighted optim.} & \includegraphics[width=\imwidth]{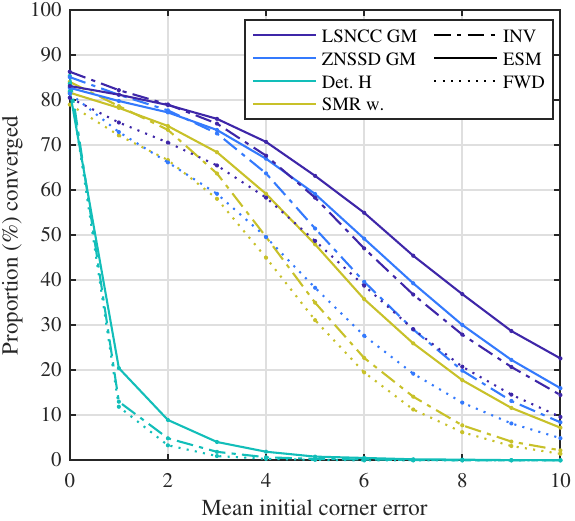} & \includegraphics[width=\imwidth]{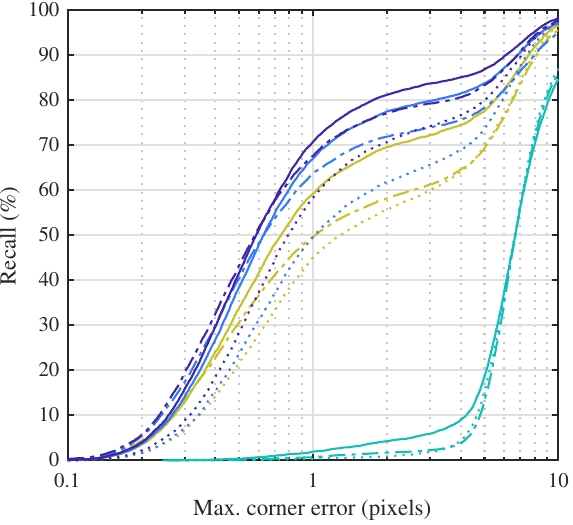} & \includegraphics[width=\imwidth]{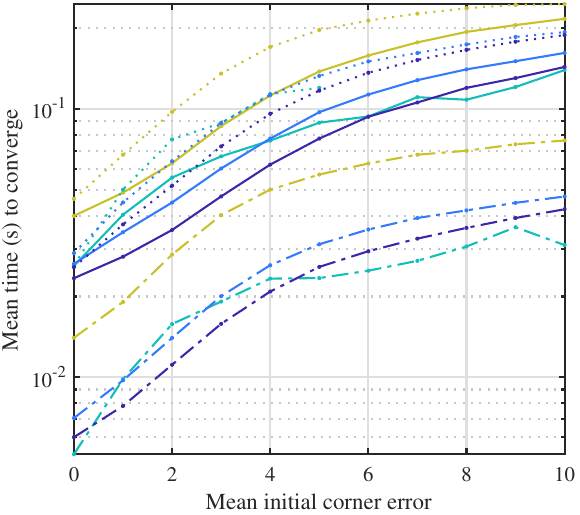} & \includegraphics[width=\imwidth]{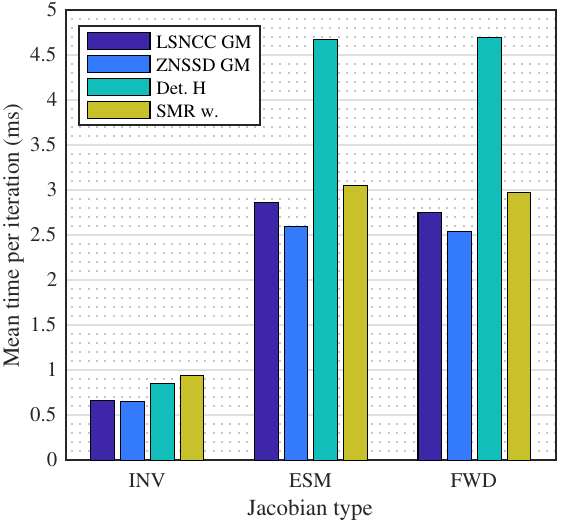} \\
		\rotatebox{90}{\footnotesize \hspace{2pt}(d) NCC \vs Image transform.} & \includegraphics[width=\imwidth]{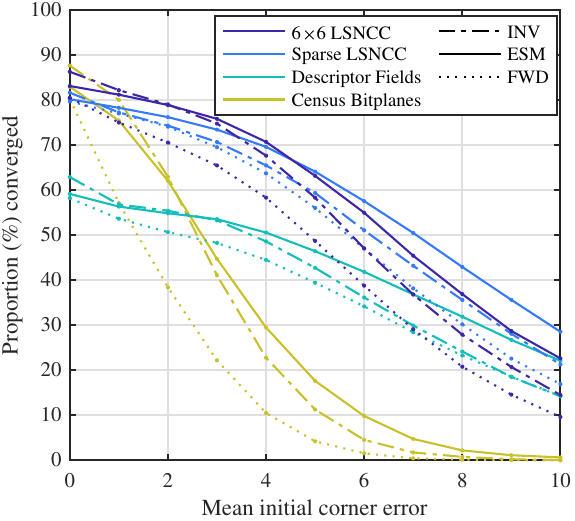} & \includegraphics[width=\imwidth]{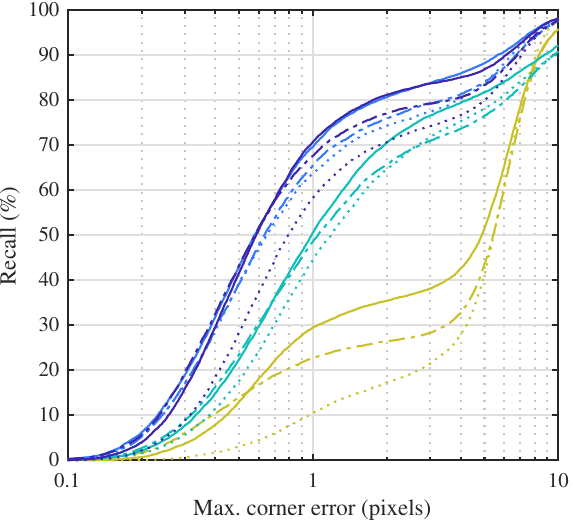} & \includegraphics[width=\imwidth]{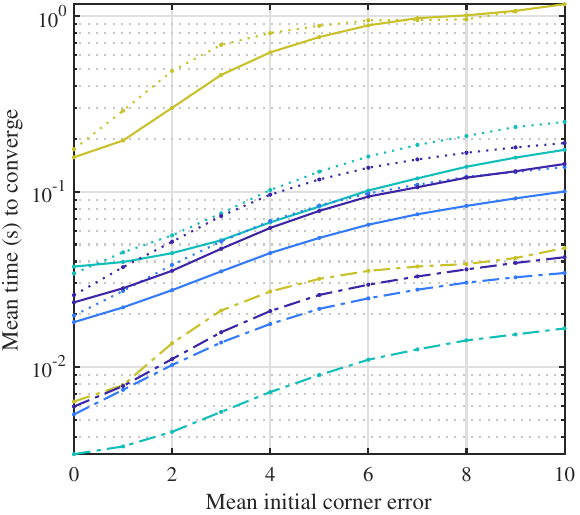} & \includegraphics[width=\imwidth]{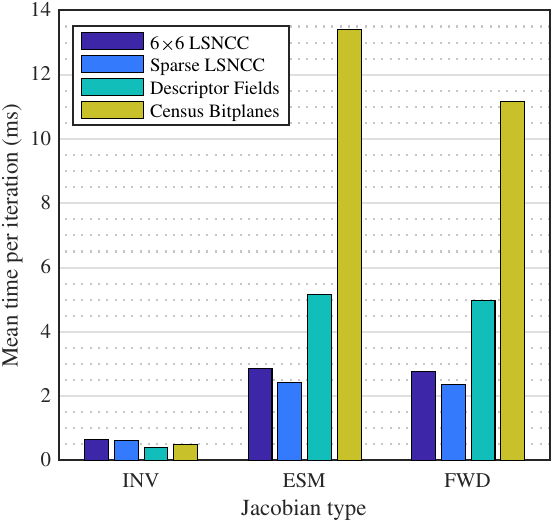} \\
		\rotatebox{90}{\footnotesize \hspace{26pt}(e) With occlusions} & \includegraphics[width=\imwidth]{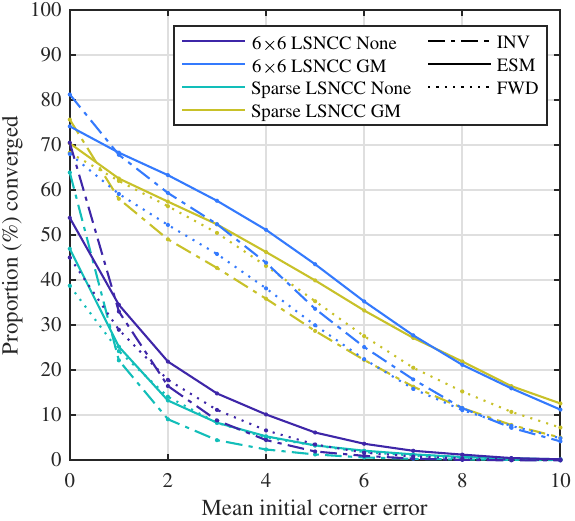} & \includegraphics[width=\imwidth]{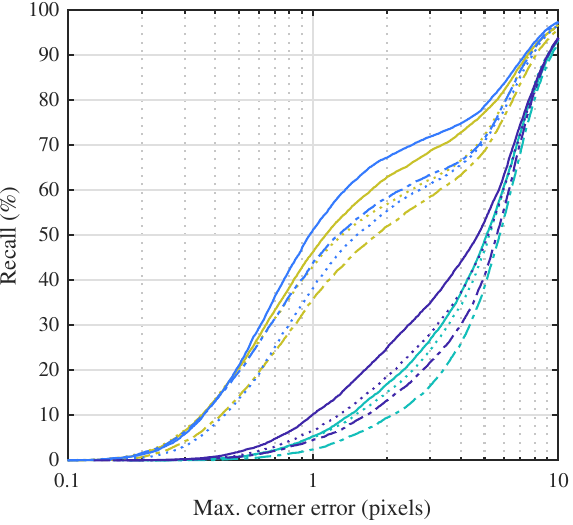} & \includegraphics[width=\imwidth]{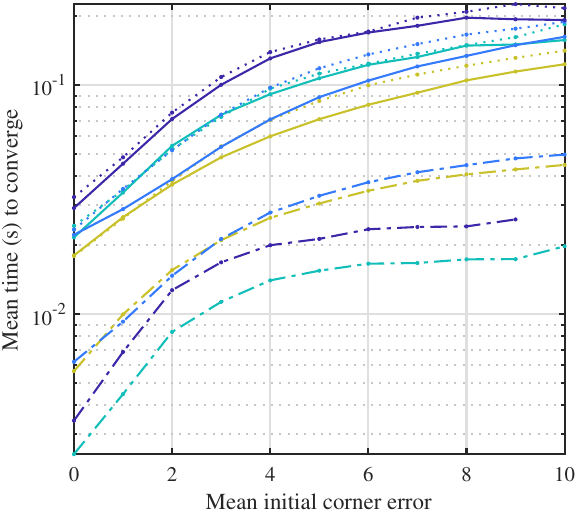} & \includegraphics[width=\imwidth]{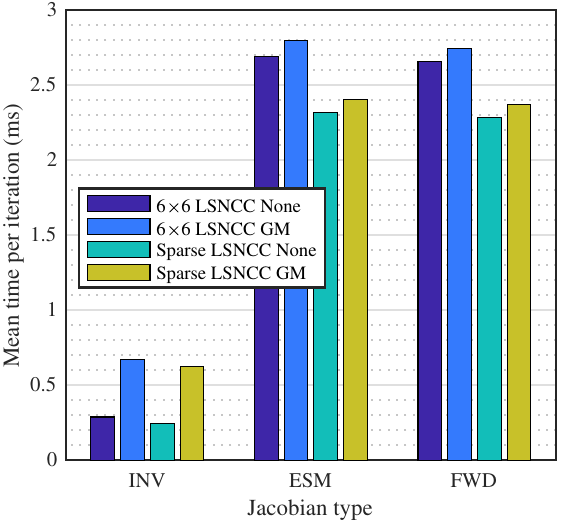} \\
		& 1. Convergence rates & 2. Recall \vs error & 3. Mean convergence time & 4. Mean iteration time
	\end{tabularx}
	\caption{{\bf Quantitative results.} Columns visualize: 1. The proportion of tests that converged to the correct solution, against the mean initial corner error (distance, in pixels). 2. Recall (\% of tests with accuracy below the threshold) against the final corner error threshold (distance, in pixels), for tests starting with a mean corner error of 4 pixels. 3. Total optimization time, averaged over all tests that successfully converged, against the mean initial corner error. 4. Time per iteration (in milliseconds), averaged over all iterations of all tests. Rows are discussed in section \ref{sec:quantitative_eval}.
	\label{fig:all_quantitative}}
\end{figure*}

\begin{figure*}
	\setlength{\imwidth}{0.24\linewidth}
	\centering
	\begin{tabularx}{\linewidth}{@{}p{10pt}@{\hspace{2pt}}*{3}{c@{\extracolsep{\fill}}}c@{}}
		\rotatebox{90}{\footnotesize \hspace{26pt}(a) Local block size} & \includegraphics[width=\imwidth]{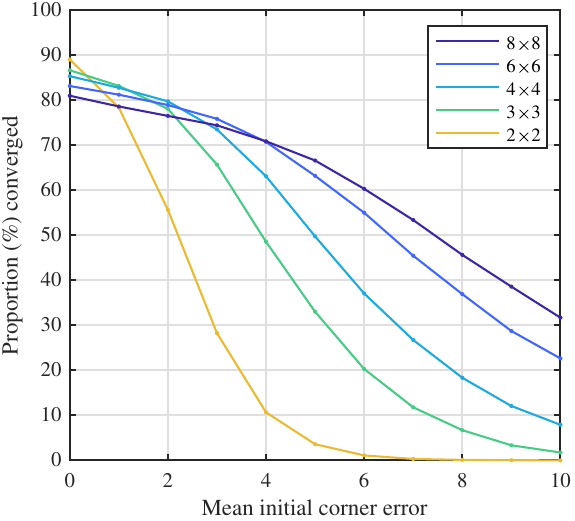} & \includegraphics[width=\imwidth]{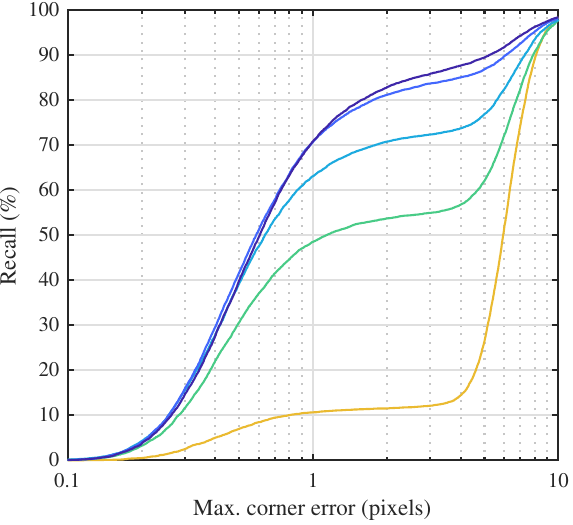} & \includegraphics[width=\imwidth]{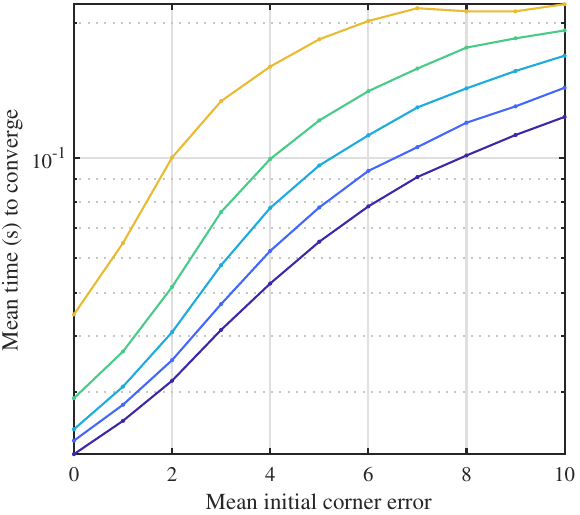} & \includegraphics[width=\imwidth]{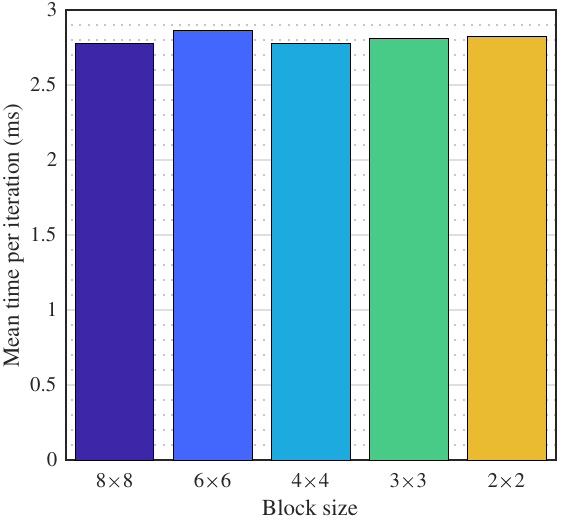} \\
		\rotatebox{90}{\footnotesize \hspace{12pt}(b) Num. sparse features} & \includegraphics[width=\imwidth]{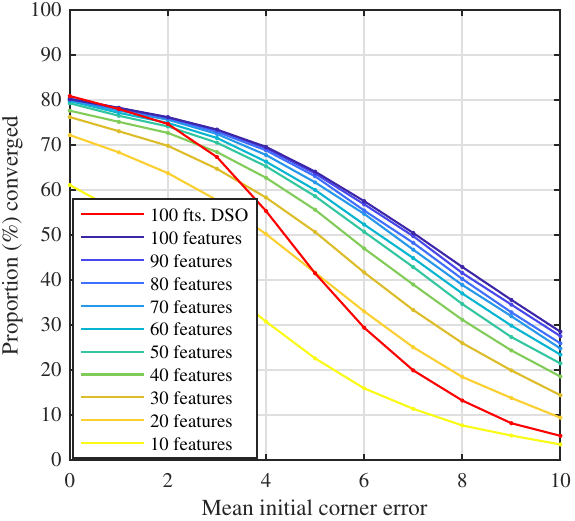} & \includegraphics[width=\imwidth]{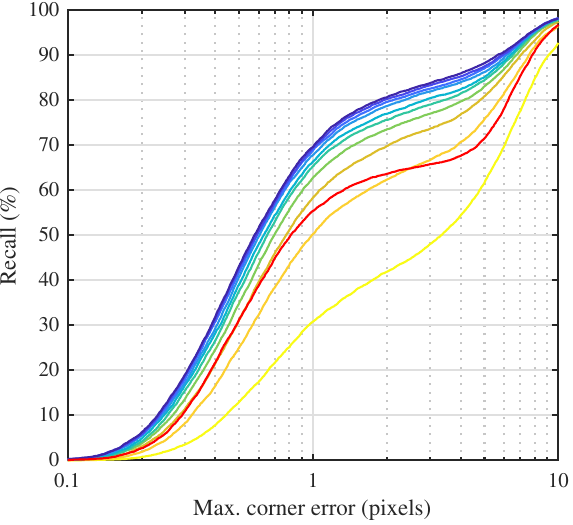} & \includegraphics[width=\imwidth]{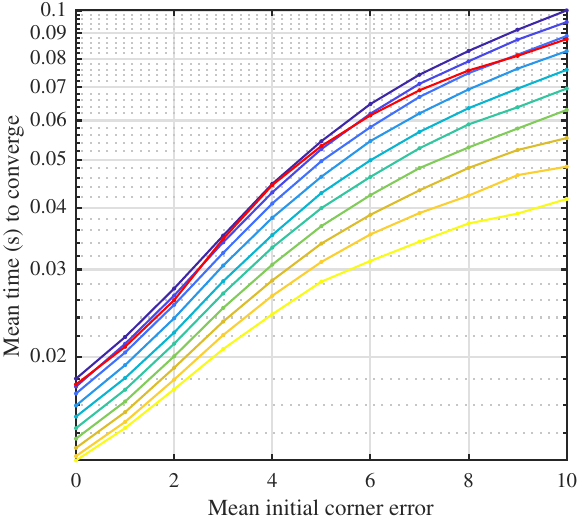} & \includegraphics[width=\imwidth]{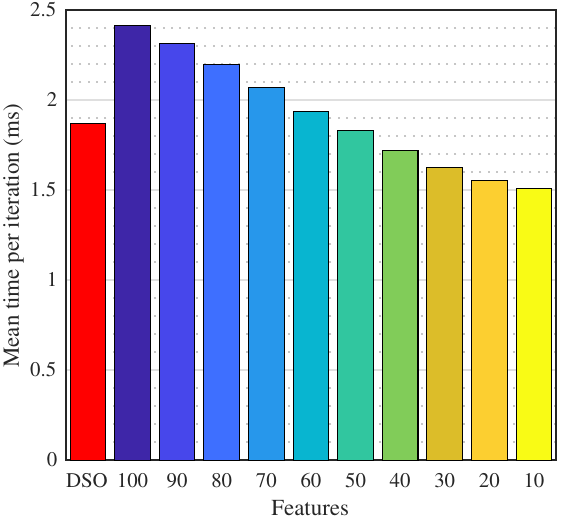} \\
		& 1. Convergence rates & 2. Recall \vs error & 3. Mean convergence time & 4. Mean iteration time
	\end{tabularx}
	\caption{{\bf Comparisons of block size, and number of sparse features.} Visualizations as per Figure~\ref{fig:all_quantitative} are given for evaluations over (a) the size of grid blocks that are locally normalized in the robust, locally normalized cost, and (b) the number of features used in the robust, sparse cost. The latter evaluation includes a result for the 8 pixel sample layout proposed in ``Direct Sparse Odometry'' (DSO)~\cite{engel2017direct}. Results are given for the ESM Jacobian only.
	\label{fig:esm_quantitative}}
\end{figure*}

Occlusions occur when a part of the target object is no longer visible, either because it goes behind another object (or even part of itself), or because it leaves the view frustrum of the camera. To understand the importance of robustifying the LSNCC cost in mitigating such occlusions, experiments are run simulating a 25\% occlusion, by replacing a random quadrant of each target test region with salt and pepper noise, each pixel being randomly set to either black or white. Results for both dense and sparse LSNCC costs, with and without Geman McClure robustification (eq.\ (\ref{eqn:geman_mcclure}); denoted GM), on this occluded dataset are shown in Figure \ref{fig:all_quantitative}(e). The results show that robustification significantly improves convergence rates in the presence of occlusion, across all costs and compositions. For example, using ESM from a starting distance of 4 pixels, the robust, dense cost converges over 50\% of the time, compared to only 10\% for the unrobustified cost. However, at over twice as slow, robustification significantly increases the per iteration computation time for INV compositions, since it precludes the use of a precomputed pseudo-inverse.

\label{sec:param_ablate}

\begin{figure}
	\setlength{\imwidth}{0.49\linewidth}
	\centering
	\begin{tabularx}{\linewidth}{@{}c@{\extracolsep{\fill}}c@{}}
		\includegraphics[width=\imwidth]{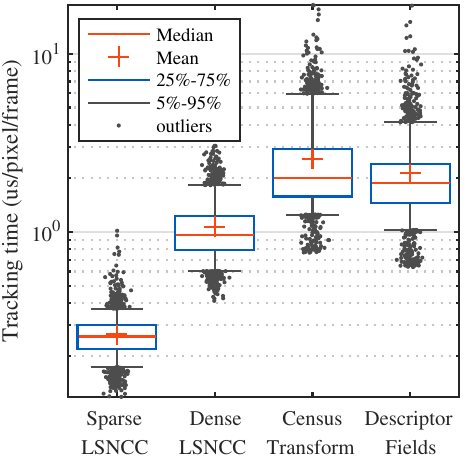} & \includegraphics[width=\imwidth]{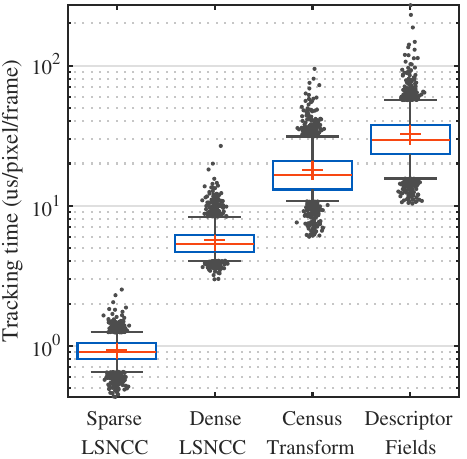} \\
		(a) INV frame times & (b) ESM frame times
	\end{tabularx}
	\caption{{\bf Video tracking times.} Distributions of per frame tracking times (in $\mu$s per tracked target pixel, per frame), normalized by the number of pixels in the reference region. \label{fig:tracking_times}}
\end{figure}


\begin{figure*}[p]
	\setlength{\imwidth}{0.485\linewidth}
	\begin{tabularx}{\linewidth}{@{}p{10pt}@{}c@{\extracolsep{\fill}}c@{}}
		\rotatebox{90}{\footnotesize \hspace{40pt}INV} & 
		\includegraphics[width=\imwidth, trim=1px 1px 1px 1px, clip]{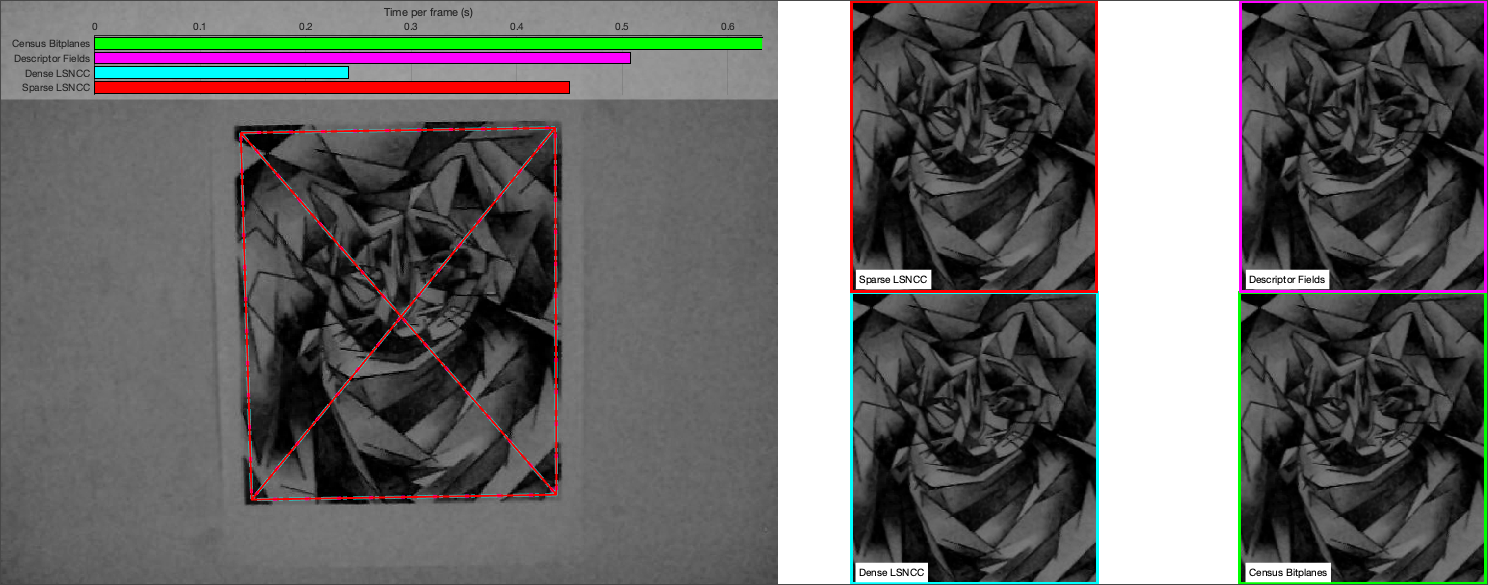} &
		\includegraphics[width=\imwidth, trim=1px 1px 1px 1px, clip]{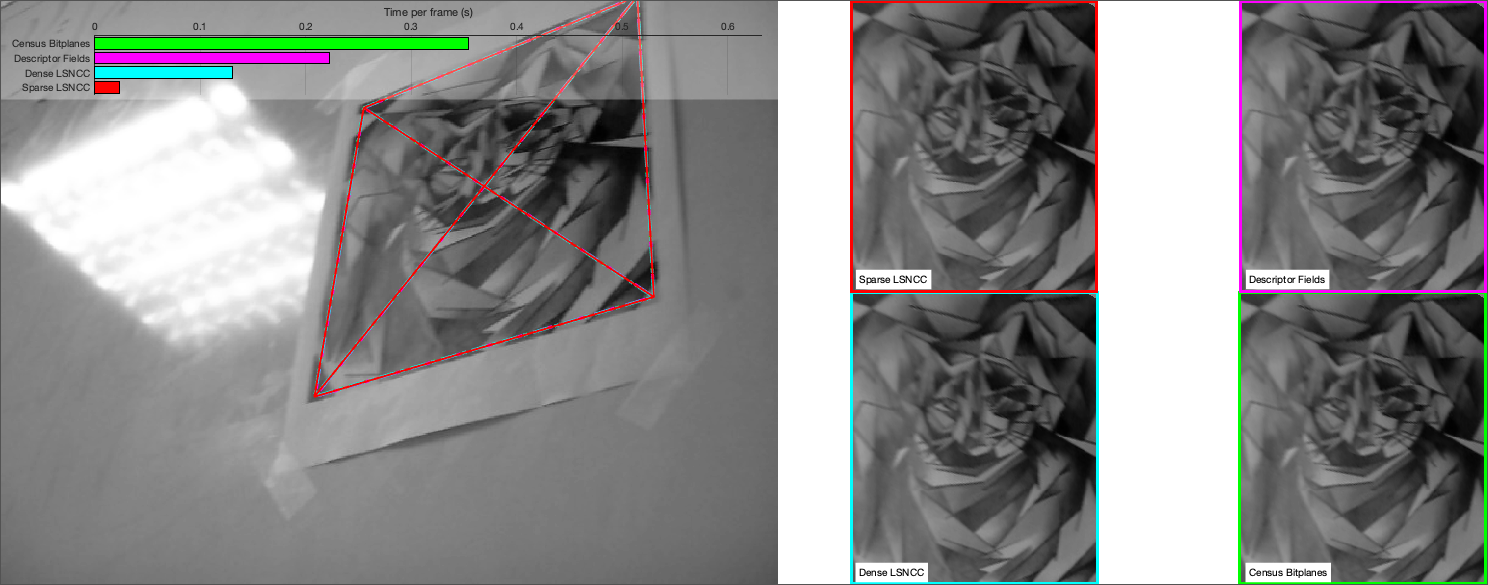}\\
		\rotatebox{90}{\footnotesize \hspace{40pt}ESM} & 
		\includegraphics[width=\imwidth, trim=1px 1px 1px 1px, clip]{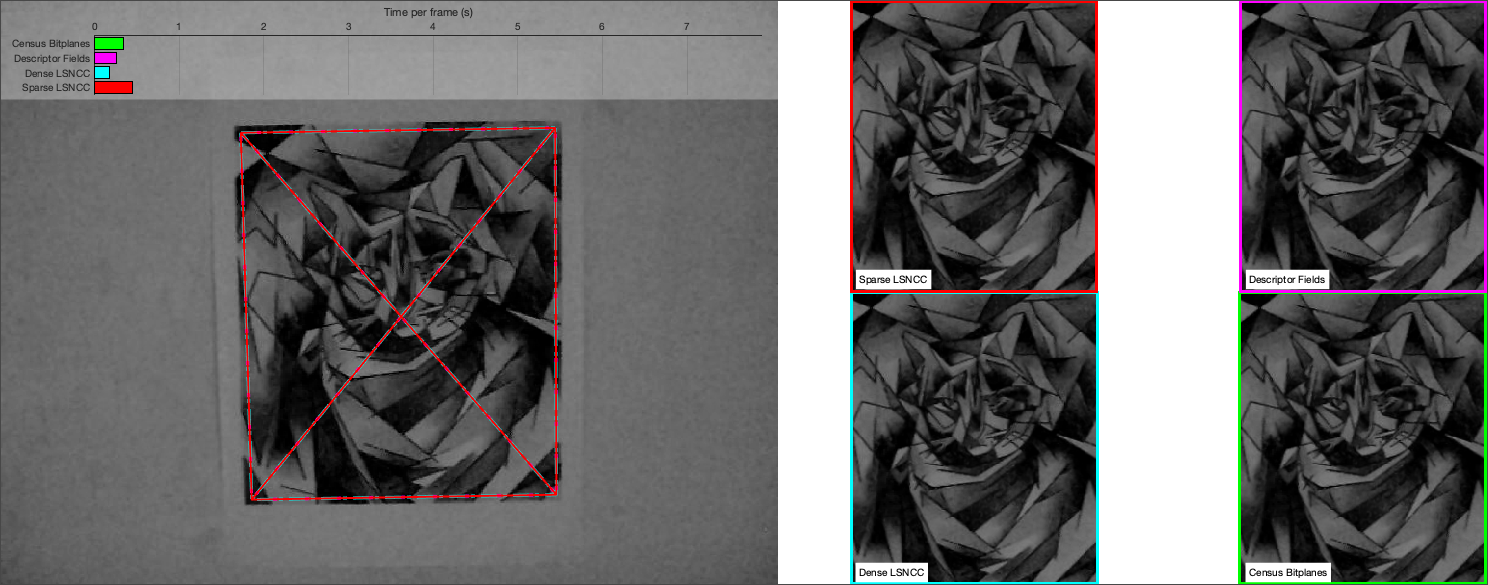} &
		\includegraphics[width=\imwidth, trim=1px 1px 1px 1px, clip]{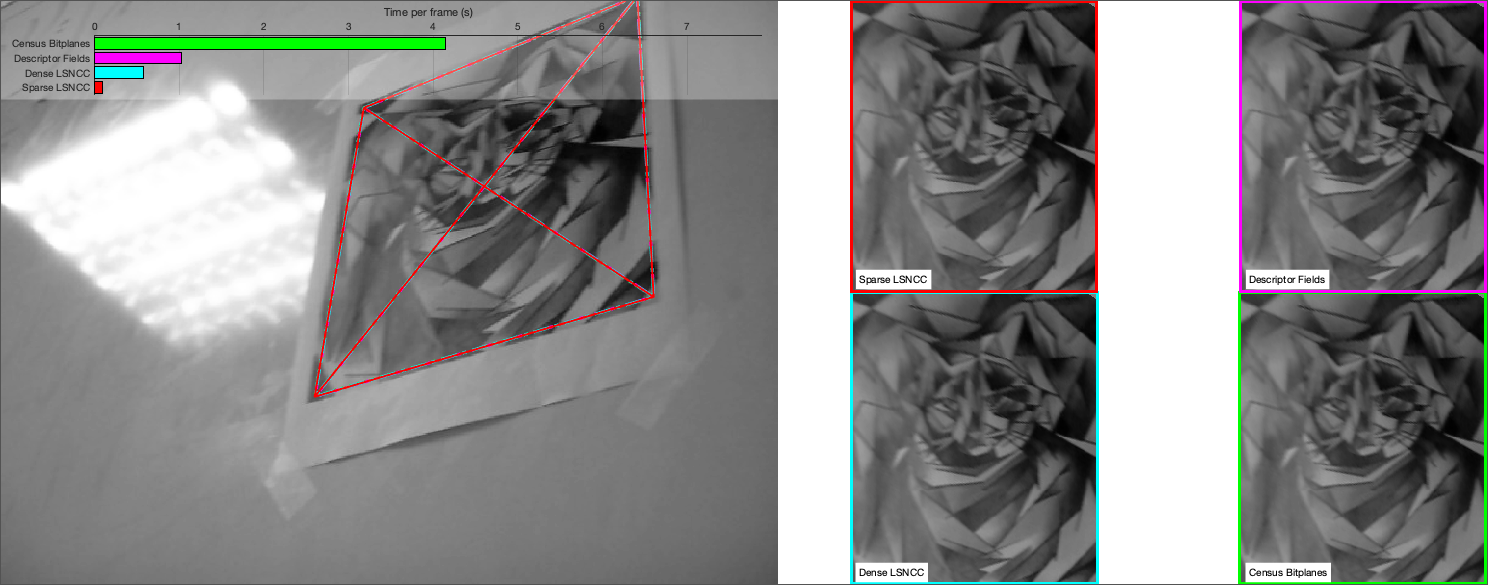}\\
		& (a) Cat Plane, frame 1 & (b) Cat plane, frame 899 (last) \\
		\rotatebox{90}{\footnotesize \hspace{40pt}INV} & 
		\includegraphics[width=\imwidth, trim=1px 1px 1px 1px, clip]{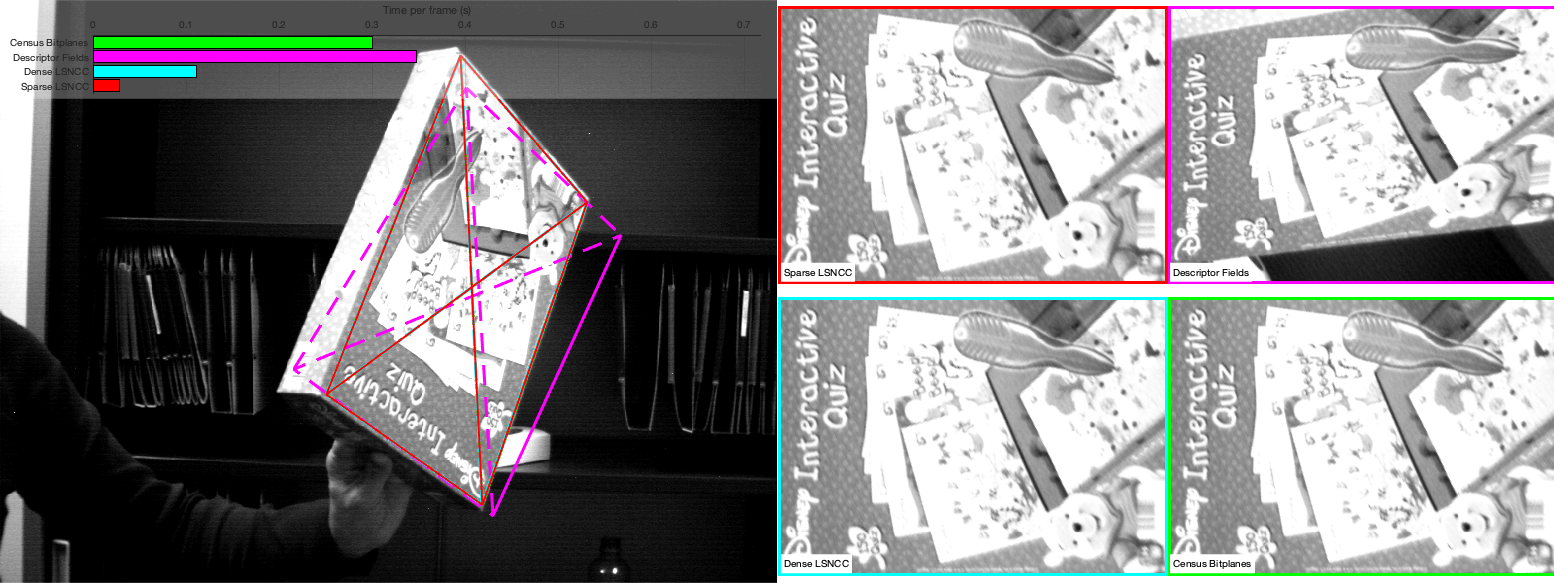}&
		\includegraphics[width=\imwidth, trim=1px 1px 1px 1px, clip]{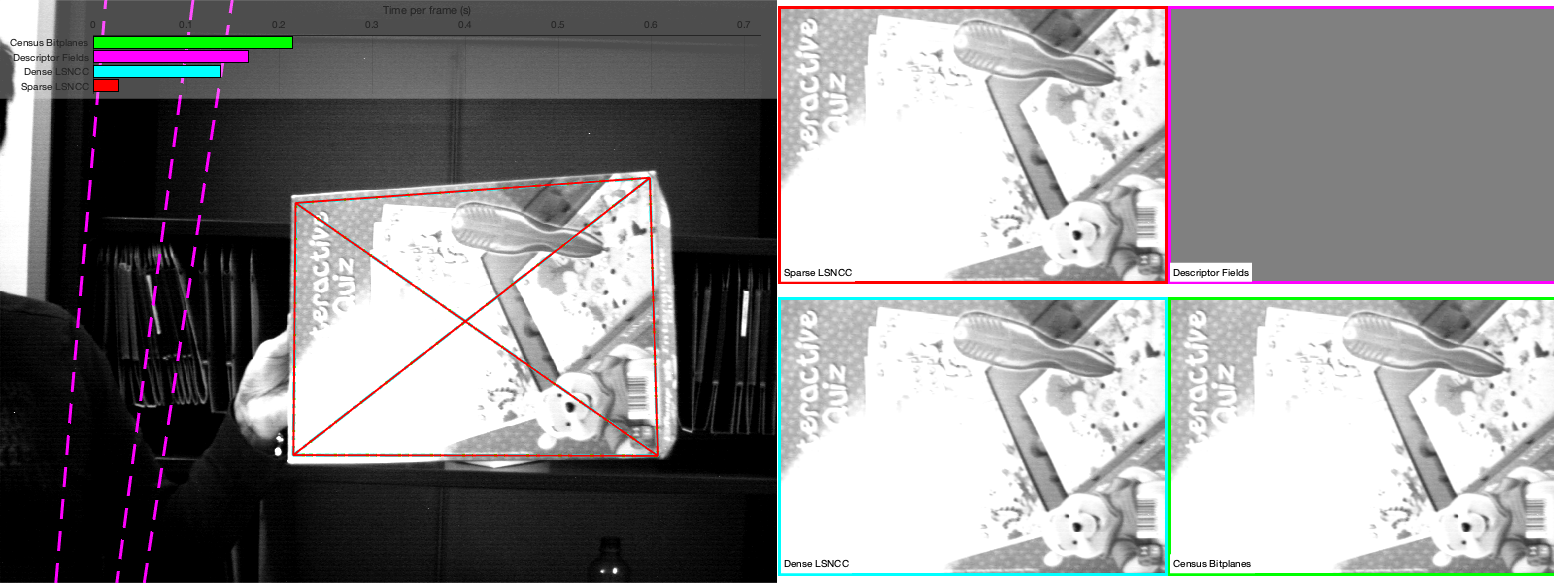}\\
		\rotatebox{90}{\footnotesize \hspace{40pt}ESM} & 
		\includegraphics[width=\imwidth, trim=1px 1px 1px 1px, clip]{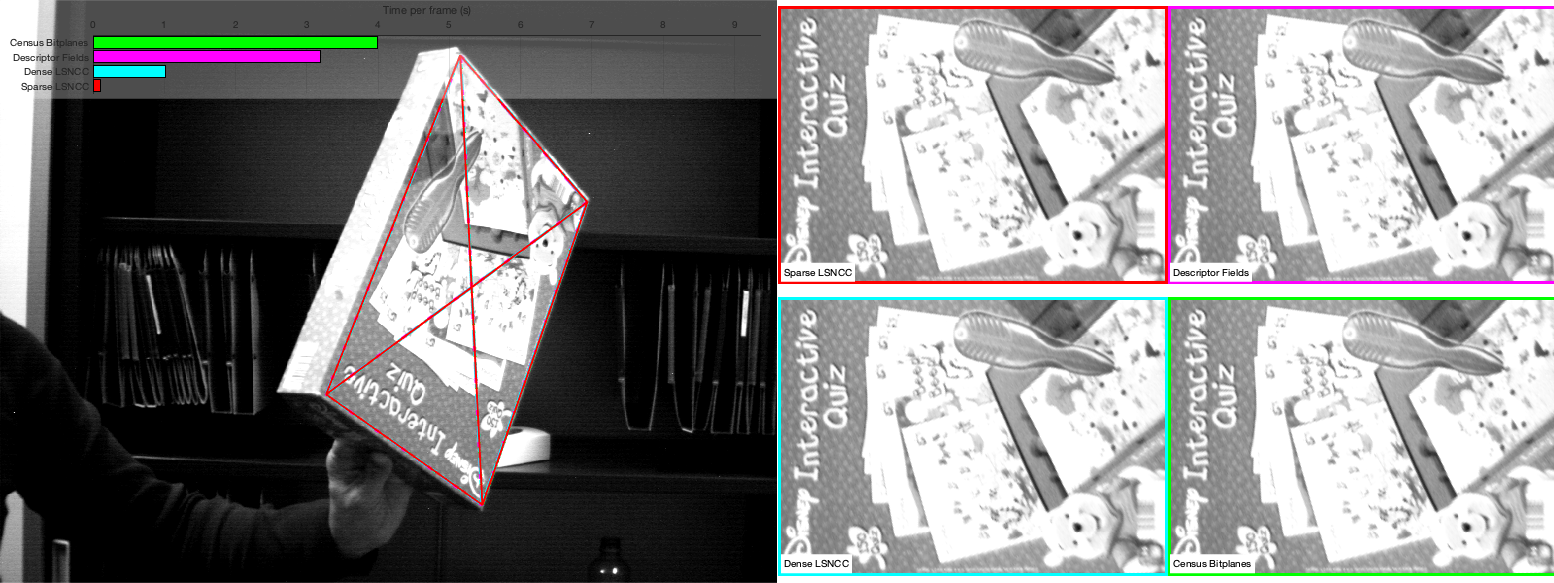}&
		\includegraphics[width=\imwidth, trim=1px 1px 1px 1px, clip]{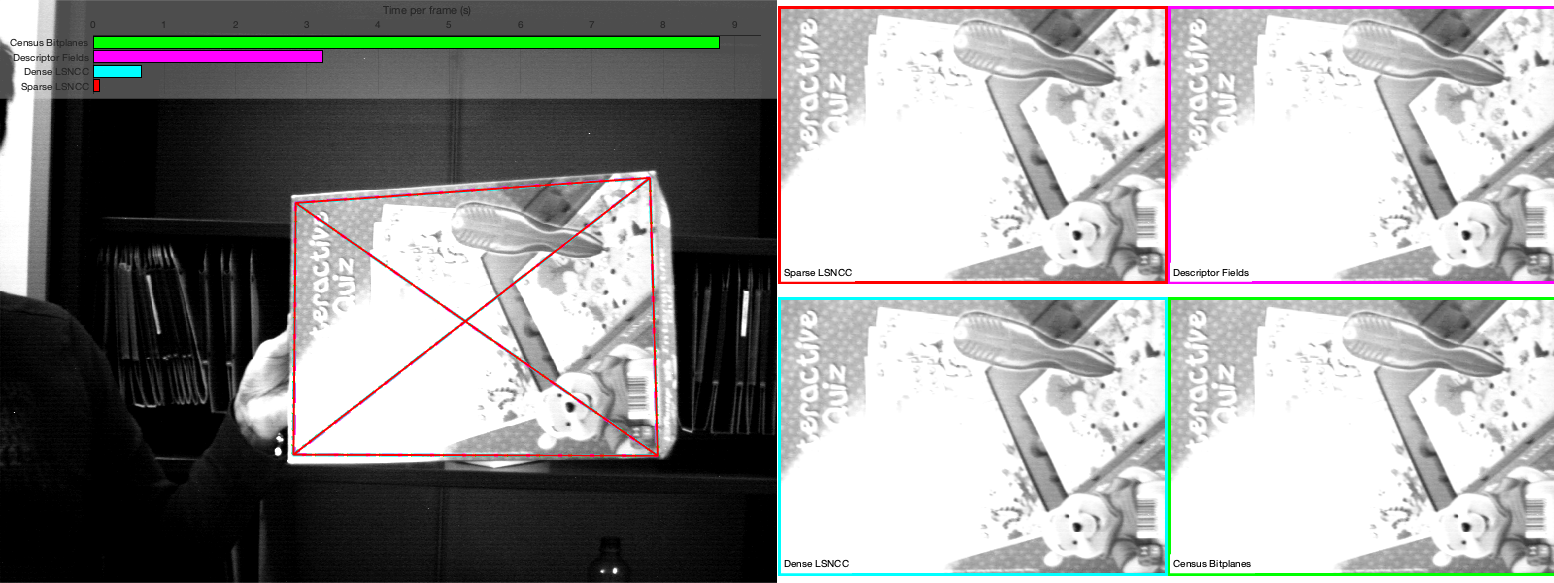}\\
		& (c) Bear, frame 216 & (d) Bear, frame 1573 (last) \\
		\rotatebox{90}{\footnotesize \hspace{40pt}INV} & 
		\includegraphics[width=\imwidth, trim=1px 1px 1px 1px, clip]{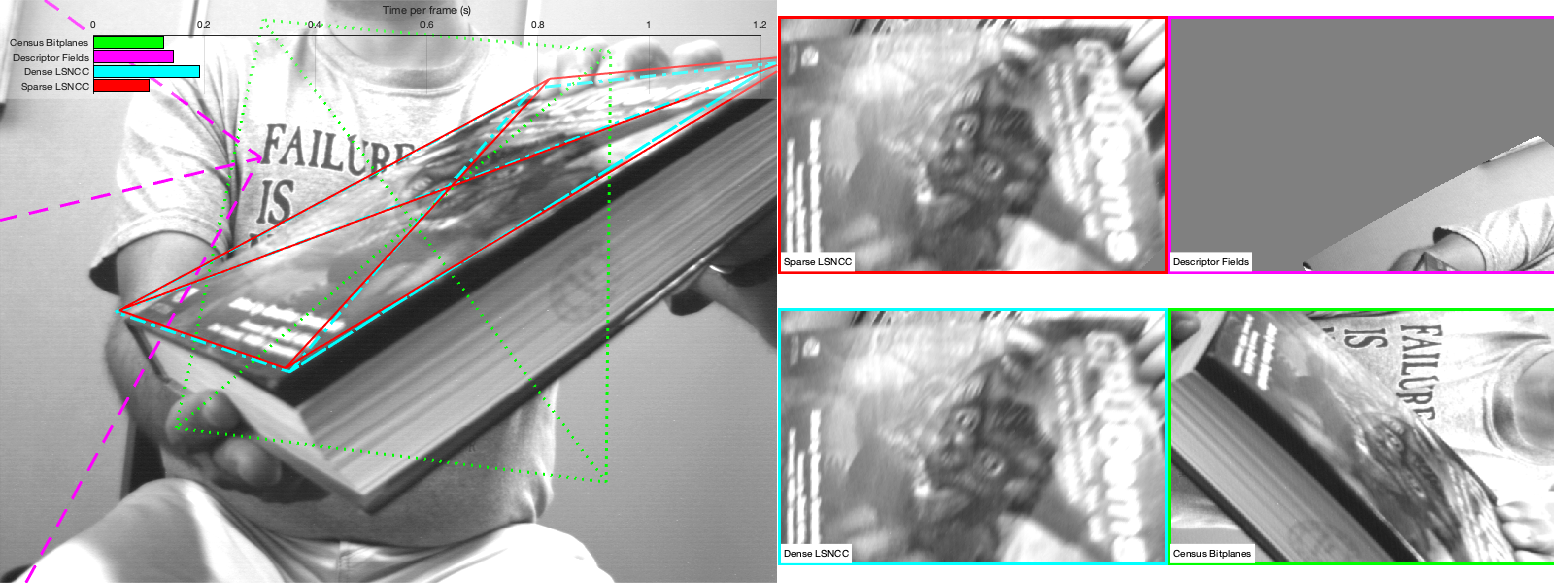}&
		\includegraphics[width=\imwidth, trim=1px 1px 1px 1px, clip]{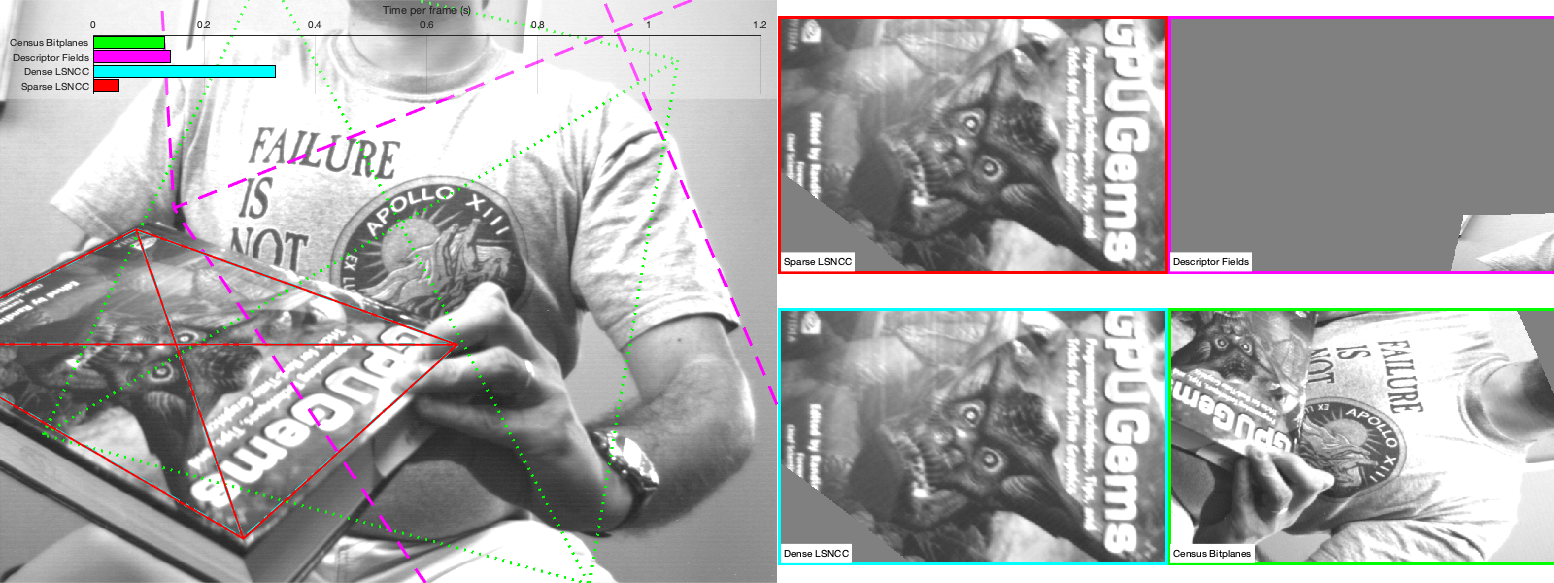}\\
		\rotatebox{90}{\footnotesize \hspace{40pt}ESM} & 
		\includegraphics[width=\imwidth, trim=1px 1px 1px 1px, clip]{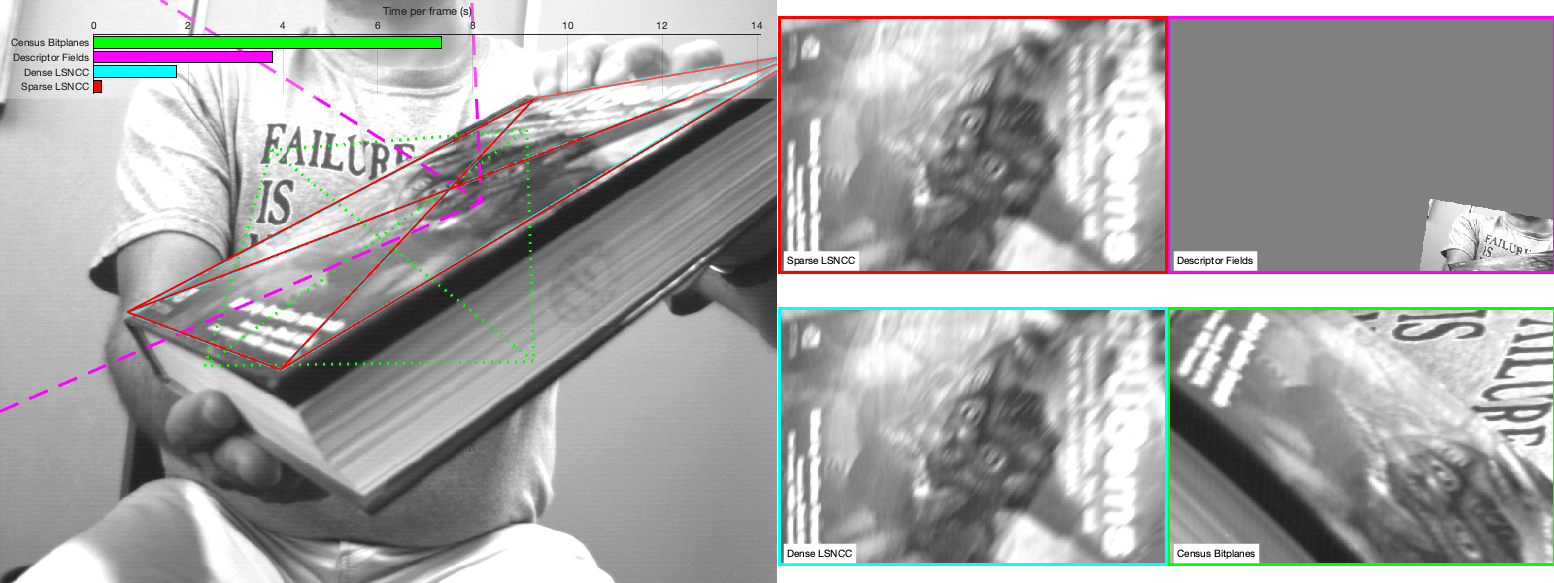}&
		\includegraphics[width=\imwidth, trim=1px 1px 1px 1px, clip]{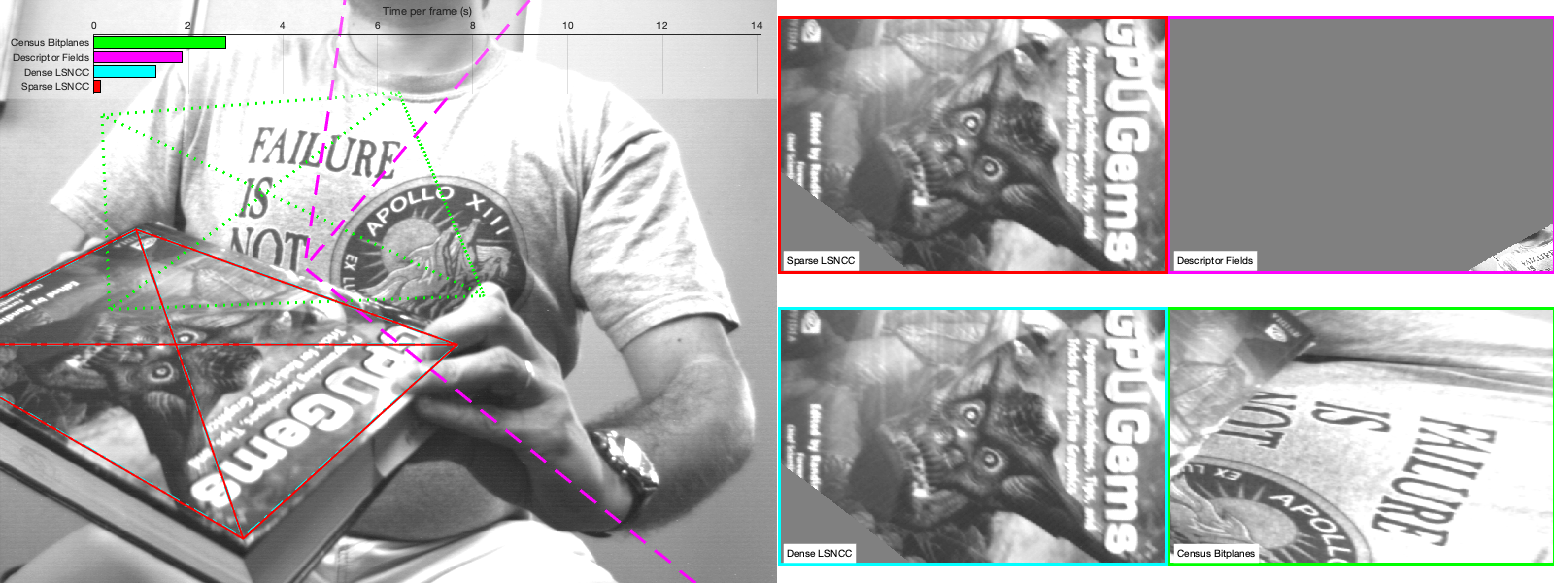}\\
		& (e) Book, frame 205 & (f) Book, frame 284 (last)
	\end{tabularx}
	\caption{{\bf Video tracking results.} Each subplot shows results for two compositions, INV (top) and ESM (bottom). The left of each result panel shows a video frame overlaid with the target rectangle estimated by each of four methods: Sparse NCC (red), Dense NCC (cyan), Descriptor Fields (magenta) and Census Bitplanes (green). The computation time for that frame, for each method, is shown in the bar chart (top). The right of each panel shows the contents of the target region for each method, warped to a canonical view.}
	\label{fig:tracking_videos}
\end{figure*}

\subsection{Tracking in videos}
\label{sec:tracking_videos}
Previous sections have presented quantitative results on a standard but contrived task, in order to show convergence statistics of different photometric costs. This section sees several methods applied to the real-world task of tracking a reference region across consecutive frames of a video. The publicly available CAT-Plane, BOOK-I and BEAR-I sequences from the works of Silveira \& Malis \cite{silveira2007real,silveira2009visual} are used, since these have strong local lighting variations, as well as occlusions by the frame boundaries.

A rectangular reference region was manually selected in the first frame of each sequence. This region was then aligned to the same region in each subsequent frame in a multi-resolution approach initialized from the homography warp computed for the previous frame (the identity warp for the first frame). Using a 5-level image pyramid constructed from each frame, the warp was optimized at the coarsest level first, progressing down to the finest. These five levels used the first 2, 4, 6, 8 \& 8 parameters of the update warp of equation (\ref{eqn:update_homog}) respectively; limiting smaller levels to lower dimensional warps made the alignment more robust.

Dense (locally normalized, 6$\times$6 blocks) and sparse (locally normalized edgelet features, 300 per pyramid level, unless dense blocks lead to fewer samples) versions of the LSNCC framework proposed here were compared to Census Bitplanes and Descriptor Fields photometric errors, for both INV and ESM compositions.\!\!\footnote{LSNCC costs are robustified only for ESM composition. Since other methods being compared are not robustified, and robustification leads to a computational penalty in the INV case (see sec.\ \ref{sec:occlusions}), not using robustification here provides a better demonstration of the accuracy \& computation time trade-offs possible using LSNCC.} A compilation of all video tracking results is available online at \texttt{\url{https://youtu.be/5GuLQXLId2c}}, with a selection of frames shown in Figure \ref{fig:tracking_videos}. Each subplot shows two panels, for INV (top) and ESM (bottom) compositions. Each panel shows (left) the video frame with coloured quadrangles overlaid indicating the region tracked by each method, as well as a bar chart (inset, top) indicating the computation time for that frame, for each method. The four rectangles (right) show a canonical, resampled version of the region tracked by each method (named bottom left; border colours correspond to the quadrangles in the video frame).

The right column of subplots (b, d \& f) show the last frame for each sequence. Tracking, once lost, is rarely reacquired. The figures show that all implementations of LSNCC successfully tracked the region all the way to the last frame. Some very fast motion on the Book sequence around frame 205 (e) causes INV LSNCC to align slightly poorly, lasting for only 3 frames; this is the only failure. In contrast, Descriptor Fields and Census Bitplanes both failed to track the Book sequence (f) using both INV and ESM compositions; this sequence contains an in-plane rotation of $>$90$^\circ$\!\!, causing these methods, which are not rotation invariant\footnote{The original implementation of INV Census Bitplanes \cite{alismail2016robust} applies the image transformation after warping, therefore it \emph{is} rotation invariant. As discussed previously, however, it doesn't support ESM.} to lose track. INV Descriptor Fields also loses track on the Bear sequence at around frame 216 (c).

In addition to reducing tracking failures, the greatest benefit from using LSNCC, and in particular the sparse cost, is the reduction in computation time, shown in Figure \ref{fig:tracking_times}. Using INV composition (a), dense LSNCC has a mean computation time around half that of the previous methods, while sparse LSNCC is almost 10 times faster than them. With ESM composition (b), the increase in speed is more marked, with dense and sparse methods 3 and 20 times faster than previous methods respectively.

\section{Conclusion}
\label{sec:conclusion}
In the roughly four decades since NCC and least squares direct methods first appeared, approaches employing NCC have been either complicated and under-used, sub-optimal, or both, whilst alternative approaches to handling local intensity variations have increased computational cost at the same time as offering lower convergence rates.

The least squares NCC framework presented here is simple, efficient and robust, finally realizing the complementary benefits of least squares optimization and the NCC measure:
\begin{itemize*}
	\item \textbf{Simple}: The least squares cost allows decades of developments in optimization and compositional photometric approaches to be easily exploited. It does not require any training, hence is broadly applicable across scenes and tasks, out of the box.
	\item \textbf{Efficient}: Single channel data, no image transformation pre-process step, an efficient Jacobian computation, an inverse compositional update (if wanted), and sparse features all reduce computation.
	\item \textbf{Robust}: The locally normalized cost offers robustness to local intensity variations, and robustifying the cost mitigates errors under partial occlusion. Furthermore, the costs are rotation invariant.
\end{itemize*}
Evaluations on aligning image regions with local variations in intensity show that this framework is more effective than both existing NCC optimizers and competing image transformation approaches at this task, as well as being significantly more efficient than the state of the art.

While tested on image alignment, this least squares NCC framework can be applied to data fields of arbitrary dimension and/or multiple channels, and extended to methods which don't just solve for alignment parameters, but also solve for other variables, such as the data coordinates $\patchcoords$. Indeed, follow-on work has already applied it to the bundle adjustment problem~\cite{woodford2020lspbm}, presenting the first such method to use a photometric cost on internet photo collections.

\appendix
\section{Equivalence to Scandaroli \etal}
\label{sec:scandaroli}
Letting $\patchJzm$, $\patchIzm$ and $\hat\jac$ denote the zero-mean source patch, target patch and Jacobian of the former respectively, Scandaroli \etal derive~\cite[eqs.~(5), (7), (8)]{scandaroli2012ncc} the following forward compositional update\footnote{Both $\M$ and $\g$ have been negated, and the latter transposed, for a simpler exposition.} for the standard NCC cost:
\begin{align}
\deltawarp &= -\M^{-1}\g,\\
\M &= \frac{\hat{\jac}}{\SDJzm}^\trsp\frac{\hat{\jac}}{\SDJzm} -\frac{\hat{\jac}}{\SDJzm}^\trsp\frac{\patchJzm}{\SDJzm}\frac{\patchJzm}{\SDJzm}^\trsp\frac{\hat{\jac}}{\SDJzm}\\
\g &= \frac{\hat{\jac}}{\SDJzm}^\trsp\left(\frac{\patchJzm}{\SDJzm}\left(\frac{\patchJzm}{\SDJzm}^\trsp\frac{\patchIzm}{\SDIzm}\right)-\frac{\patchIzm}{\SDIzm}\right)
\end{align}
$\M$ can be shown to be equal to $\overrightarrow{\jac}^\trsp\overrightarrow{\jac}$, using the variance normalization Jacobian from \eqnref{jac_ncc}, as follows:
\begin{align}
\overrightarrow{\jac}^\trsp\overrightarrow{\jac} &= \frac{\hat{\jac}}{\SDJzm}^\trsp\left(\identity - \frac{\patchJzm}{\SDJzm}\frac{\patchJzm}{\SDJzm}^\trsp\right)^\trsp\left(\identity - \frac{\patchJzm}{\SDJzm}\frac{\patchJzm}{\SDJzm}^\trsp\right) \frac{\hat{\jac}}{\SDJzm},\\
&= \frac{\hat{\jac}}{\SDJzm}^\trsp\Bigg(\identity -2 \frac{\patchJzm}{\SDJzm}\frac{\patchJzm}{\SDJzm}^\trsp+\frac{\patchJzm}{\SDJzm}\overbrace{\cancel{\frac{\patchJzm}{\SDJzm}^\trsp\frac{\patchJzm}{\SDJzm}}}^{=1}\frac{\patchJzm}{\SDJzm}^\trsp\Bigg) \frac{\hat{\jac}}{\SDJzm},\\
&= \frac{\hat{\jac}}{\SDJzm}^\trsp\Bigg(\identity - \frac{\patchJzm}{\SDJzm}\frac{\patchJzm}{\SDJzm}^\trsp\Bigg) \frac{\hat{\jac}}{\SDJzm} = \M.
\end{align}
Similarly, $\g$ can be shown to be equal to $\overrightarrow{\jac}^\trsp\error$, as follows:
\begin{align}
\overrightarrow{\jac}^\trsp\error &= \frac{\hat{\jac}}{\SDJzm}^\trsp\Bigg(\identity - \frac{\patchJzm}{\SDJzm}\frac{\patchJzm}{\SDJzm}^\trsp\Bigg)^\trsp\left(\frac{\patchJzm}{\SDJzm}-\frac{\patchIzm}{\SDIzm}\right),\\
&= \frac{\hat{\jac}}{\SDJzm}^\trsp\Bigg(\frac{\patchJzm}{\SDJzm}-\frac{\patchIzm}{\SDIzm} - \frac{\patchJzm}{\SDJzm}\overbrace{\cancel{\frac{\patchJzm}{\SDJzm}^\trsp\frac{\patchJzm}{\SDJzm}}}^{=1}+\frac{\patchJzm}{\SDJzm}\frac{\patchJzm}{\SDJzm}^\trsp\frac{\patchIzm}{\SDIzm}\Bigg)\\
&= \frac{\hat{\jac}}{\SDJzm}^\trsp\Bigg(\cancel{\frac{\patchJzm}{\SDJzm} - \frac{\patchJzm}{\SDJzm}}+\frac{\patchJzm}{\SDJzm}\frac{\patchJzm}{\SDJzm}^\trsp\frac{\patchIzm}{\SDIzm}-\frac{\patchIzm}{\SDIzm}\Bigg) = \g.
\end{align}
Thus this forward compositional update is mathematically equivalent to the least squares one presented here, computed using equation (\ref{eqn:normal_equations}). Similar derivations can be applied to their inverse compositional update, demonstrating equivalence also. However, their proposed combined update~\cite[eq. (13)]{scandaroli2012ncc} differs from the ESM formulation, since
\begin{align}
\textrm{SMR:}&&\deltawarp &= -\left(\overrightarrow{\M}+\overleftarrow{\M}\right)^{-1}\left(\overrightarrow{\g}^\trsp+\overleftarrow{\g}^\trsp\right),\\
&&&= -\left(\overrightarrow{\jac}^\trsp\overrightarrow{\jac}+\overleftarrow{\jac}^\trsp\overleftarrow{\jac}\right)^{-1}(\overrightarrow{\jac}^\trsp\error+\overleftarrow{\jac}^\trsp\error),\\
&&&= -\left(4\overleftrightarrow{\jac}^\trsp\overleftrightarrow{\jac}-2\overrightarrow{\jac}^\trsp\overleftarrow{\jac}\right)^{-1}2\overleftrightarrow{\jac}^\trsp\error,\\
\textrm{ESM:}&&& \ne -\left(\overleftrightarrow{\jac}^\trsp\overleftrightarrow{\jac}\right)^{-1}\overleftrightarrow{\jac}^\trsp\error.
\end{align}
The improvements in convergence rate and computation time using standard least squares ESM alignment, shown in Figure \ref{fig:scandaroli_esm}, and the improvements due to using a standard robustifier over an ad-hoc weighting scheme, shown in Figure \ref{fig:all_quantitative}(c), both underscore the benefits of casting NCC in a least squares form (and deriving the resulting Jacobian) over the bespoke implementation of Scandaroli \etal. Doing so confers considerable advantages in ease of implementation, as well as access to multiple, proven enhancements from a substantial body of literature.



\ifCLASSOPTIONcompsoc
\else
\fi

\ifCLASSOPTIONcaptionsoff
\newpage
\fi

{
	\bibliographystyle{IEEEtran}
	\bibliography{references}
}

\begin{IEEEbiography}[{\includegraphics[width=1in,height=1.25in,clip,keepaspectratio]{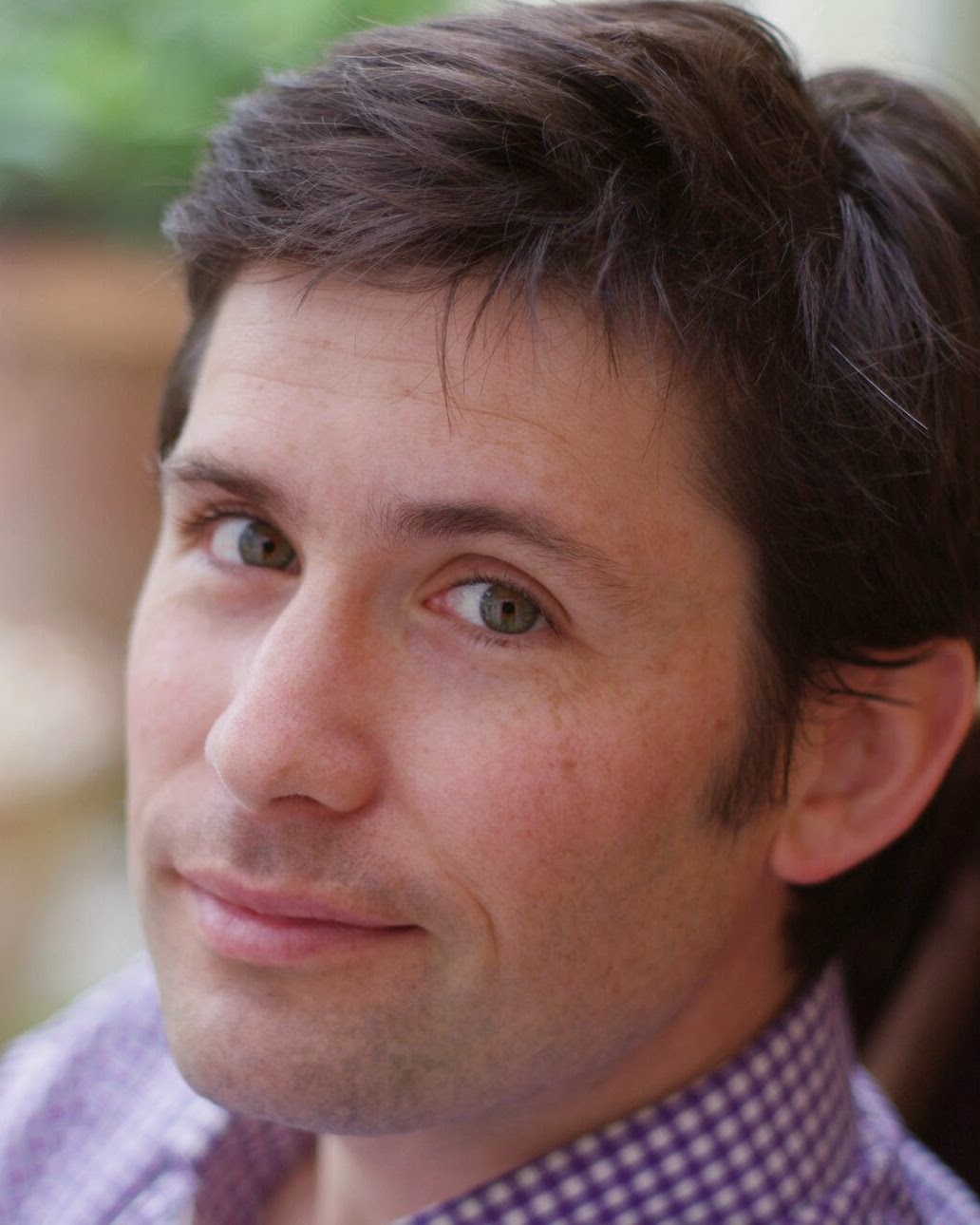}}]{Oliver J. Woodford}
This work was done while Oliver Woodford was a Lead Research Scientist at Snap Inc, where he worked from Nov.\ 2016 to Dec.\ 2020. His research encompasses the areas of 3D reconstruction and optimization, and has won awards including an IEEE CVPR best paper award, and a Toshiba Research \& Development Achievement Award whilst a Research Engineer at Toshiba Research Europe's laboratory in Cambridge (2009-2014). He has also worked for several technology startups. He received the MEng degree in Engineering from the University of Cambridge in 2002, and a DPhil in Engineering from the University of Oxford in 2009, under the supervision of Andrew Fitzgibbon, Philip Torr and Ian Reid.
\end{IEEEbiography}
	
\end{document}